\documentclass{article}

 \usepackage[eandd, final]{neurips_2026}
\usepackage[utf8]{inputenc} % allow utf-8 input
\usepackage[T1]{fontenc}    % use 8-bit T1 fonts
\usepackage{hyperref}       % hyperlinks
\usepackage{url}            % simple URL typesetting
\usepackage{booktabs}       % professional-quality tables
\usepackage{amsfonts}       % blackboard math symbols
\usepackage{nicefrac}       % compact symbols for 1/2, etc.
\usepackage{microtype}      % microtypography
\usepackage{xcolor}         % colors
\usepackage{subcaption}
\usepackage{graphicx}
\usepackage{multirow}

\usepackage{amsmath}
\usepackage{amssymb}% http://ctan.org/pkg/amssymb
\usepackage{pifont}% http://ctan.org/pkg/pifont
\newcommand{\cmark}{\ding{51}}%
\newcommand{\xmark}{\ding{55}}%
\usepackage[table]{xcolor}
\usepackage{makecell}
\usepackage{stackengine}
\usepackage{tikz,pgfplots,pgfplotstable}
\usepackage{amsmath}
\pgfplotsset{compat=default}
\usetikzlibrary{pgfplots.groupplots}

\usetikzlibrary{tikzmark}

\pgfplotsset{compat=1.11,
    /pgfplots/ybar legend/.style={
    /pgfplots/legend image code/.code={%
       \draw[##1,/tikz/.cd,yshift=-0.25em]
        (0cm,0cm) rectangle (3pt,0.8em);},
   }, 
}  

\title{PixFoundation 2.0: Do Video Multi-Modal LLMs Use Motion in Visual Grounding?}

\author{%
  Mennatullah Siam\\
  Ro'ya Community\thanks{Ro'ya is a community that I founded for Africans working on Responsible AI and CV. It is not registered as an organization; yet our events and research have existed since 2022.  \url{https://ro-ya-cv4africa.github.io/homepage}. The use of the first-person plural in this text is strictly for double-blind review. \textbf{e-mail:} mennatul@ualberta.ca}\\
}

\begin{document}

\maketitle

\begin{abstract}

Multi-modal large language models (MLLMs) have shown impressive generalization across tasks using images and text modalities. While their extension to video has enabled tasks such as video question answering and video captioning, their pixel-level visual grounding abilities are less studied. In this work, we raise the pertinent question of whether motion is used in pixel-level visual grounding and whether video MLLMs can segment objects based on natural language expressions describing their motion patterns. We identify the shortcomings in the current benchmarks, where we show that a single frame can often suffice for capturing the referring expression without any temporal reasoning. To address this, we introduce novel motion-centric probing, particularly designed for the visual grounding task, to study video MLLMs' ability to identify true motion from a fake one and their ability to grasp the ordering of motion. Consequently, we introduce MoCentric-Bench, a motion-centric benchmark designed to evaluate video MLLMs based on their ability to capture the interaction between motion and language, rather than relying primarily on static cues. We further establish strong single-image baselines that either match or surpass prior methods. Finally, we explore a simple motion-centric adaptation that provides state-of-the-art performance on our MoCentric-Bench. Code is available at \url{https://msiam.github.io/PixFoundationSeries/}. 

%Code and datasets will be made publicly available as part of the PixFoundation Series~\url{https://msiam.github.io/PixFoundationSeries/}.% at \url{https://github.com/MSiam/PixFoundation-2.0.git}.
\end{abstract}

\section{Introduction} % 1.5 page

\begin{figure}[t]
\centering
\includegraphics[width=\textwidth]{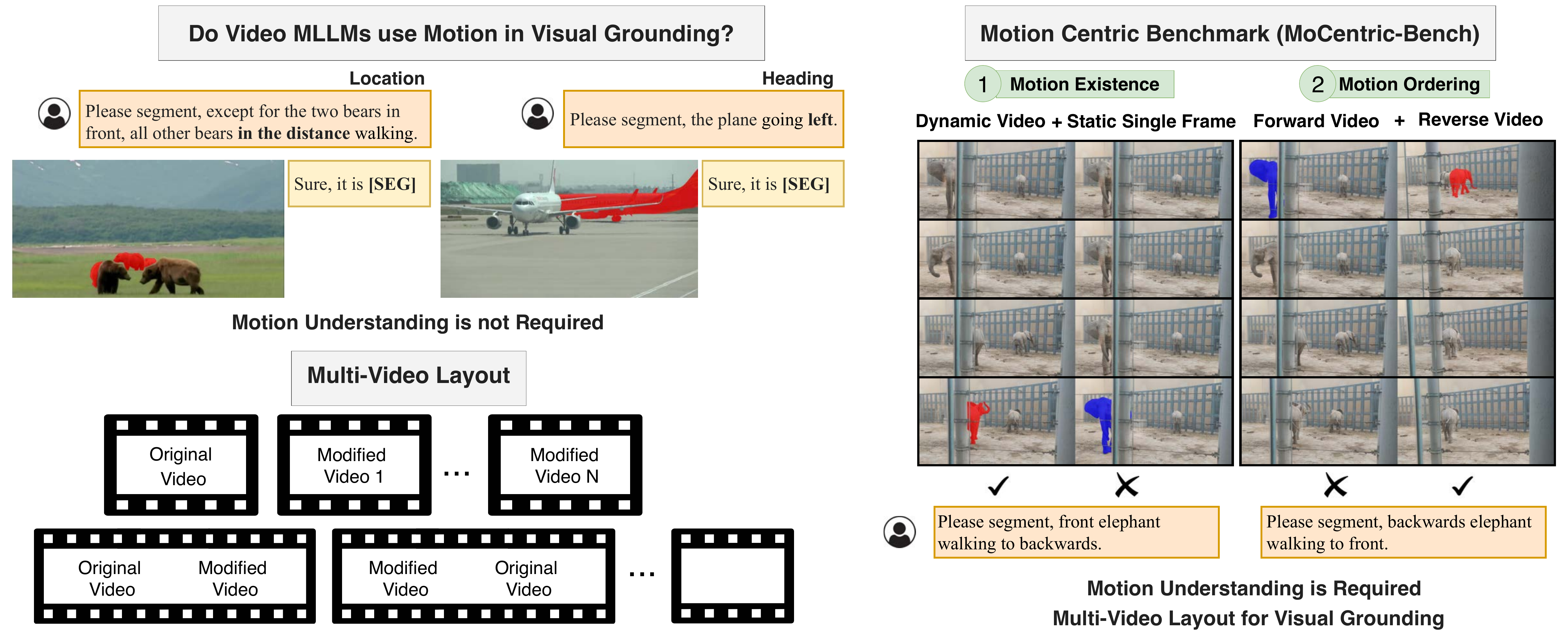}
\vspace{-1em}
\caption{\textbf{Left:} \textit{(Top)} Shortcomings of the current referring segmentation benchmarks, where the expression can be captured from a single frame. \textit{(Bottom)} Illustration of the multi-video layout used in our benchmark as a probe to quantify failures of video MLLMs in capturing motion. \textbf{Right:} Our MoCentric-Bench evaluates the use of motion in visual grounding within videos. Our probing is designed for visual grounding using a static single frame or the reverse of the video. The correct and deceiving ground-truth segmentations are highlighted in red and blue, respectively.} 
\label{fig:overview}
\vspace{-2em}
\end{figure}

Multi-modal large language models (MLLMs) have recently emerged as general-purpose tools that can operate on input image/video and text~\cite{liu2023visual,bai2023qwen,bai2025qwen2,liu2024improved,zhu2025internvl3}. They can be language guided through instructions in addition to various visual prompting techniques to produce the desired output. The extension of multi-modal large language models to operate on videos has been extensively studied~\cite{maaz2023video,fu2024video,bai2025qwen2,zohar2024apollo,llin2023videoo}. Nonetheless, these mainly focused on video question answering or video captioning. However, few methods focused on \textit{dense spatiotemporal tasks with MLLMs}, such as pixel-level visual grounding in videos that include referring video segmentation and grounded conversation generation~\cite{yan2024visa,munasinghe2024videoglamm,yuan2025sa2va}.\\ %Pixel-level MLLMs are generally designed to incorporate segmentation as part of their output to visually ground their output, whether these models are handling images or videos~\cite{lai2024lisa, zhang2024omg, siam2025pixfoundation}. 
%Video segmentation generally focuses on identifying different segments in a video that can be defined based on semantics, saliency or language guided~\cite{zhou2022survey}. The latter is the main focus of this work, where models aim to segment objects of interest in videos based on a referring expression. 

Arguably, visual grounding can be conducted on the bounding box level without resorting to the more challenging pixel-level segmentation task~\cite{bai2025qwen2,zhu2025internvl3}. Nonetheless, we believe that pixel-level tasks are a stronger direction, even if proven more challenging, because of their interplay with other crucial pixel-level ones. These include depth estimation, motion estimation, tracking and prediction, all of which are closely tied to developing powerful and interpretable general-purpose embodied intelligence~\cite{szot2025multimodal}. Various works emerged in studying pixel-level visual grounding in videos~\cite{yan2024visa,munasinghe2024videoglamm,yuan2025sa2va,clark2026molmo2}, along with the respective benchmarks pushing towards better understanding and reasoning capabilities~\cite{yan2024visa,ding2023mevis,siam2025pixfoundation}. Previous work~\cite{siam2025pixfoundation} focused on \textit{vision-centric} benchmarking for pixel-level visual grounding and visual question answering using a paired evaluation. The goal of the study is to ensure that the emerging research is in the direction of improving pixel-level MLLMs' abilities without degrading their fundamental language capabilities. 

Inspired by such direction, we focus on \textit{motion-centric} benchmarking. We ask the major question: \textit{``Do video MLLMs use motion in visual grounding?''}. Various interpretability and benchmarking efforts emerged to question the developed video understanding models' ability to capture motion information and dynamics~\cite{kowal2022deeper,kowal2024quantifying,buch2022revisiting,xue2025seeing,zohar2024apollo}. However, these methods were either confined to precursors of MLLMs without language interaction or focused on simpler tasks such as video question answering, refer to Table~\ref{tab:probing}. It is crucial to highlight that a proper language and motion interaction in video MLLMs entails understanding both local motion and higher-level dynamics. As such, we aim to ensure video MLLMs do not use shortcuts to localize motion patterns based on a single frame or a few static keyframes, but we rather aim for a proper understanding of both levels with emphasis on local motion. We propose a novel probing that is designed specifically for visual grounding in short to medium-length videos, while leaving long-form video understanding outside the scope of this work. \textit{Dense spatiotemporal segmentation in videos is an interesting test-bed for studying whether video MLLMs capture motion and its relationship with language. It also allows us to assess whether these models rely on appearance cues from single or a few static frames instead of learning genuine local motion. Importantly, visual grounding is central to embodied intelligence, where understanding local motion is critical for safety.} Concurrent work~\cite{clark2026molmo2}, released after ours, focused on tracking, yet this does not necessarily reflect a proper language and motion interaction with an understanding of motion patterns in relation to language. Hence, our study is crucial to quantify failures in this interaction and is orthogonal to theirs.

\begin{table}[t]
\caption{\textbf{Video understanding studies}, with their respective probing methods. Our work is the first to study visual grounding in videos. Probing w/ input indicates a controlled change in the input to the model towards assessing certain ability or bias. Rep.: representations, OF: Optical Flow, CM: caption matching, DC: direction classification.}
\resizebox{0.95\textwidth}{!}{
\centering
\begin{tabular}{l|c|p{45mm}|p{30mm}}
\toprule
Method & Year & Probing w/ input & Task  \\ \toprule
RESOUND~\cite{li2018resound} & ECCV 2018 & Object, Scene, People Rep.  & Action Recog. \\ 
Scene Bias~\cite{choi2019can} & CVPR 2019 & Mask Out & Action Recog., Temp. Loc., Action Det.\\ 
Dyn.Stat.~\cite{kowal2022deeper} & CVPR 2022 & Shuffle, Perturb OF & Action Recog., Video Seg. w/o Language  \\ 
ATP~\cite{buch2022revisiting} & CVPR 2022 & Single Frame Rep. & Video QA, Video Ret.  \\ \midrule
Apollo~\cite{zohar2024apollo} & CVPR 2025 & Single Frame, Text Only & Video QA \\
TV-Bench~\cite{cores2024tvbench} & Arxiv 2024 & Single Frame, Shuffle, Reverse, Text Only & Video QA  \\ 
AOT~\cite{xue2025seeing} & NeurIPS 2025 & Reverse, Shuffle &  Video QA, CM, DC. \\ \midrule
\rowcolor{gray!50} MoCentric-Bench (ours) & NeurIPS 2026 & Single Frame, \newline Reverse,  \newline Multi-Video Layout w/ Reverse, \newline Multi-Video Layout w/ Single Frame & Visual Grounding \\  \bottomrule
\end{tabular}}
\vspace{-1em}
\label{tab:probing}
\end{table}

We provide MoCentric-Bench, a motion-centric benchmark and focus on two aspects: (i) identifying fake motion from a single frame vs. true motion in the video, and (ii) understanding the ordering of motion and differentiating expressions describing the original video vs. its reverse. Our attempt highlights the problems in the current benchmarks and methods with empirical means. Figure~\ref{fig:overview} presents the motivation of our MoCentric-Bench. \underline{In summary, our contributions are threefold:}
(i) we conduct an empirical analysis revealing the limitations of current video MLLM benchmarks and models in distinguishing real from fake motion and in capturing the ordering of motion within visual grounding;
(ii) we introduce novel motion-centric probing, which is used to synthesize datasets for both training and evaluation; and
(iii) we establish strong single-image baselines and a simple motion-centric adaptation that surpasses prior methods.

\section{Related Work} % 1.5 page
\noindent \textbf{Multi-modal large language models (MLLMs) and benchmarking.} %rewrite
Pioneering works in multi-modal large language models such as LLaVA~\cite{liu2024visual,liu2024improved}, Molmo~\cite{molmo}, Qwen-VL~\cite{bai2023qwen} and InternVL~\cite{chen2024internvl} have driven significant development towards the creation of general-purpose models. Consequent works built upon these developments to equip MLLMs with better spatial and temporal understanding emerged~\cite{bai2025qwen2,chen2024expanding,zhu2025internvl3,lai2024lisa,zhang2025llava,munasinghe2024videoglamm,yan2024visa,zohar2024apollo}.  Some of these methods can perform visual grounding in either images or videos in terms of bounding boxes, points or masks~\cite{lai2024lisa,zhang2025llava,molmo,munasinghe2024videoglamm,yan2024visa}. 
%Other works focused on extending to video MLLMs~\cite{bai2025qwen2,munasinghe2024videoglamm,yan2024visa,zohar2024apollo}.
One of the major drivers behind these developments is the evaluation benchmarks that push the limit on these models and ensure improved performance, in addition to studies that interpret their behaviour. There is an abundance of standard benchmarks used to evaluate MLLMs~\cite{kazemzadeh2014referitgame,fang2024mmbench,fu2024video}. Concurrent work studying video MLLMs~\cite{zohar2024apollo} has shown the bias within such evaluation benchmarks towards using a single image or textual input only instead of fully evaluating the use of temporal information. Nonetheless, the majority of previous works on video MLLMs benchmarking focused on simpler tasks, e.g, video question answering.  Our work focuses on the visual grounding task with carefully designed probing techniques for it. %While there are recent benchmarks evaluating pixel-level visual grounding in videos~\cite{yan2024visa,ding2023mevis}, we show that they are ineffective in assessing the ability of video MLLMs to capture motion. As such, we propose novel probing techniques that are motion-centric and designed specifically for the visual grounding task.

\noindent \textbf{Video segmentation.} The general task of video segmentation takes as input a video clip and outputs segments within the video based on the definition for the objects of interest that can either be: (i) based on semantic categories (i.e., video semantic segmentation)~\cite{miao2021vspw}, (ii) within foreground/background segmentation framework relying on saliency or tracking (i.e., video object segmentation)~\cite{karim2023med}, or (iii) based on language (i.e., referring video segmentation)~\cite{munasinghe2024videoglamm,yan2024visa,wu2022language,ding2023mevis}. We focus on referring video segmentation that relies on a referring expression describing the object/s of interest to be segmented within an input video. One track of methods focused on emphasizing motion in referring video segmentation~\cite{ding2023mevis}. However, our work shows the shortcomings in the aforementioned benchmarks, and we show that state-of-the-art methods and our proposed baselines that surpass them fail in our motion-centric evaluation.

\section{Method} % 2 page
\begin{figure*}[t]
\centering
\begin{subfigure}{0.2\textwidth}
\includegraphics[width=\textwidth]{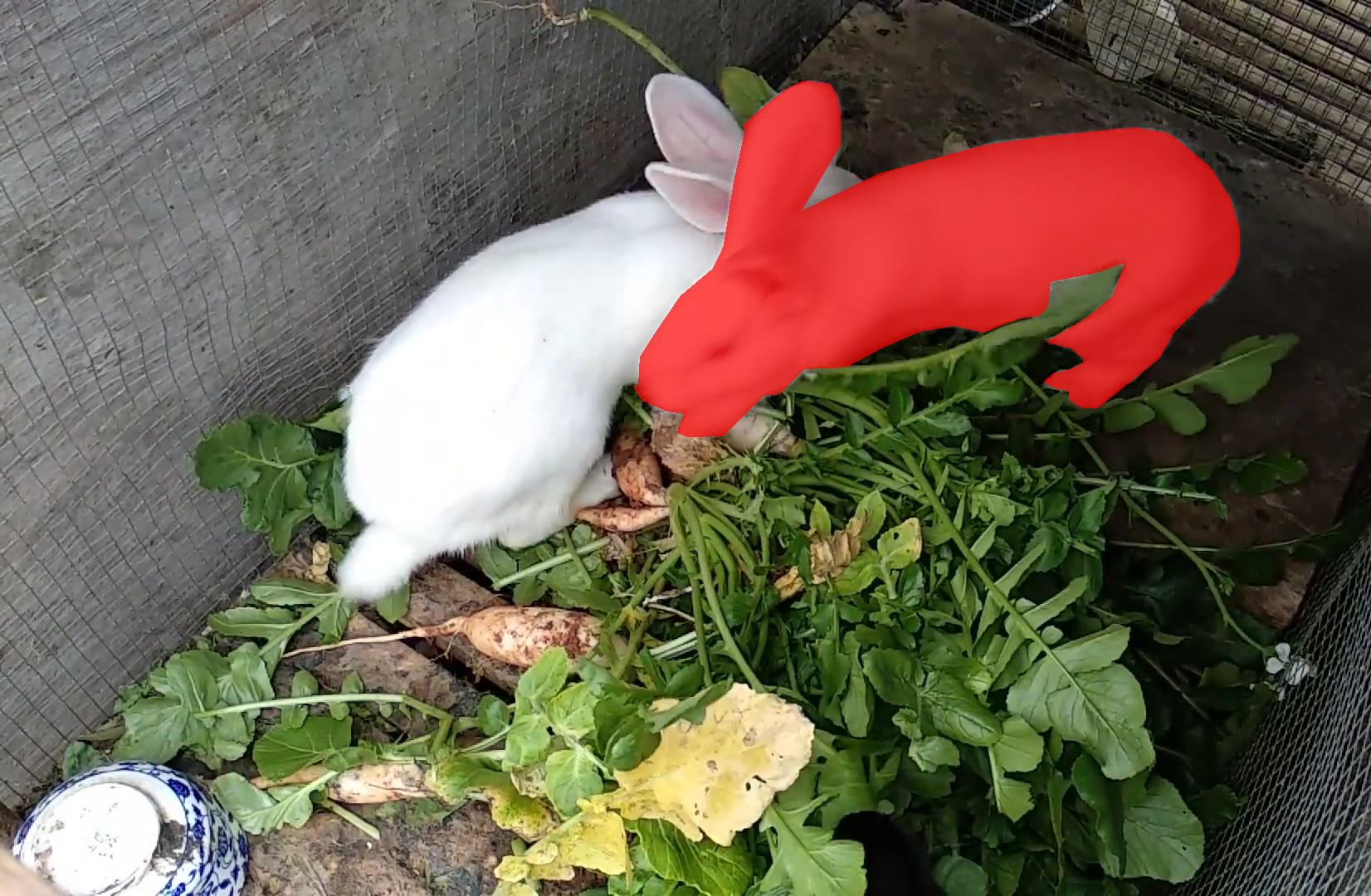}
\caption{}
\label{fig:keyframes-a}
\end{subfigure}%
\begin{subfigure}{0.062\textwidth}
\includegraphics[width=\textwidth]{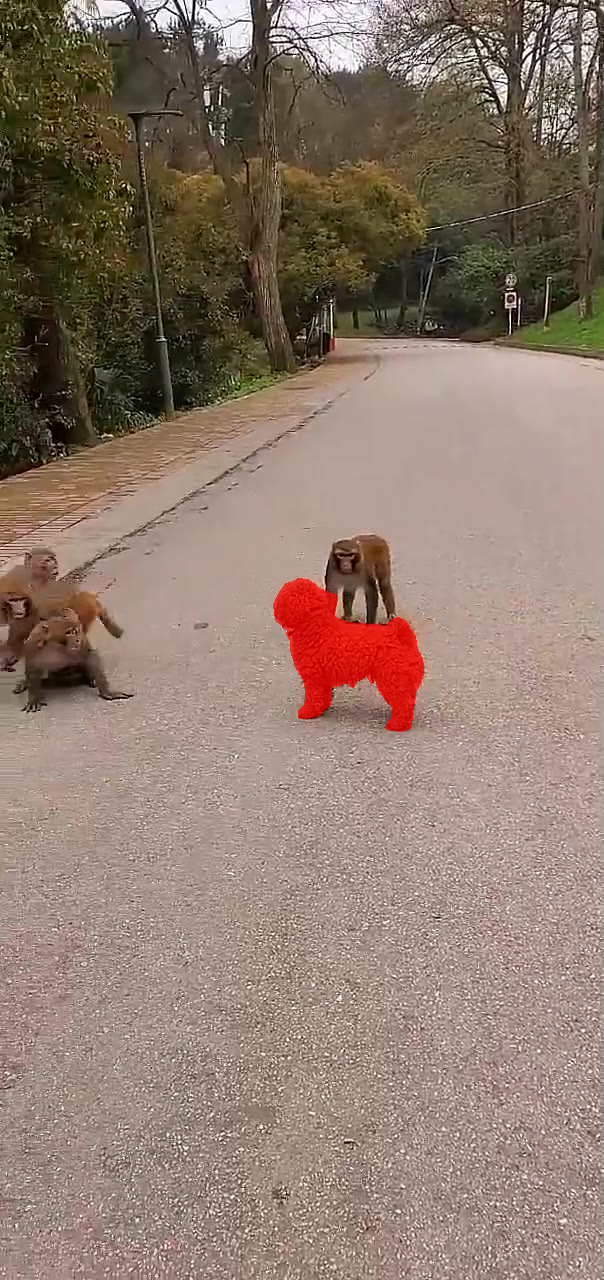}
\caption{}
\label{fig:keyframes-b}
\end{subfigure}%
\begin{subfigure}{0.175\textwidth}
\includegraphics[width=\textwidth]{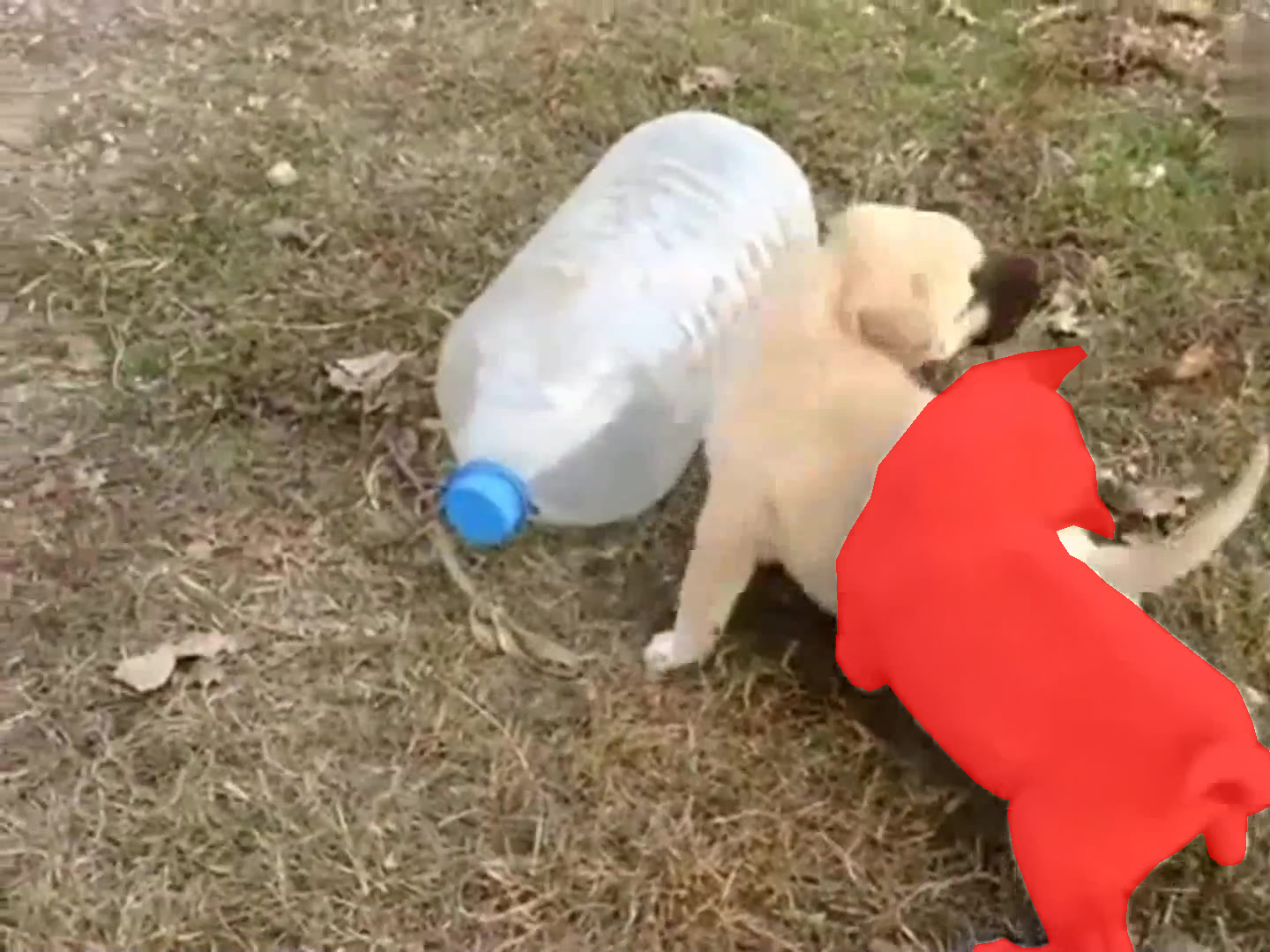}
\caption{}
\label{fig:keyframes-c}
\end{subfigure}%
\begin{subfigure}{0.282\textwidth}
\includegraphics[width=\textwidth]{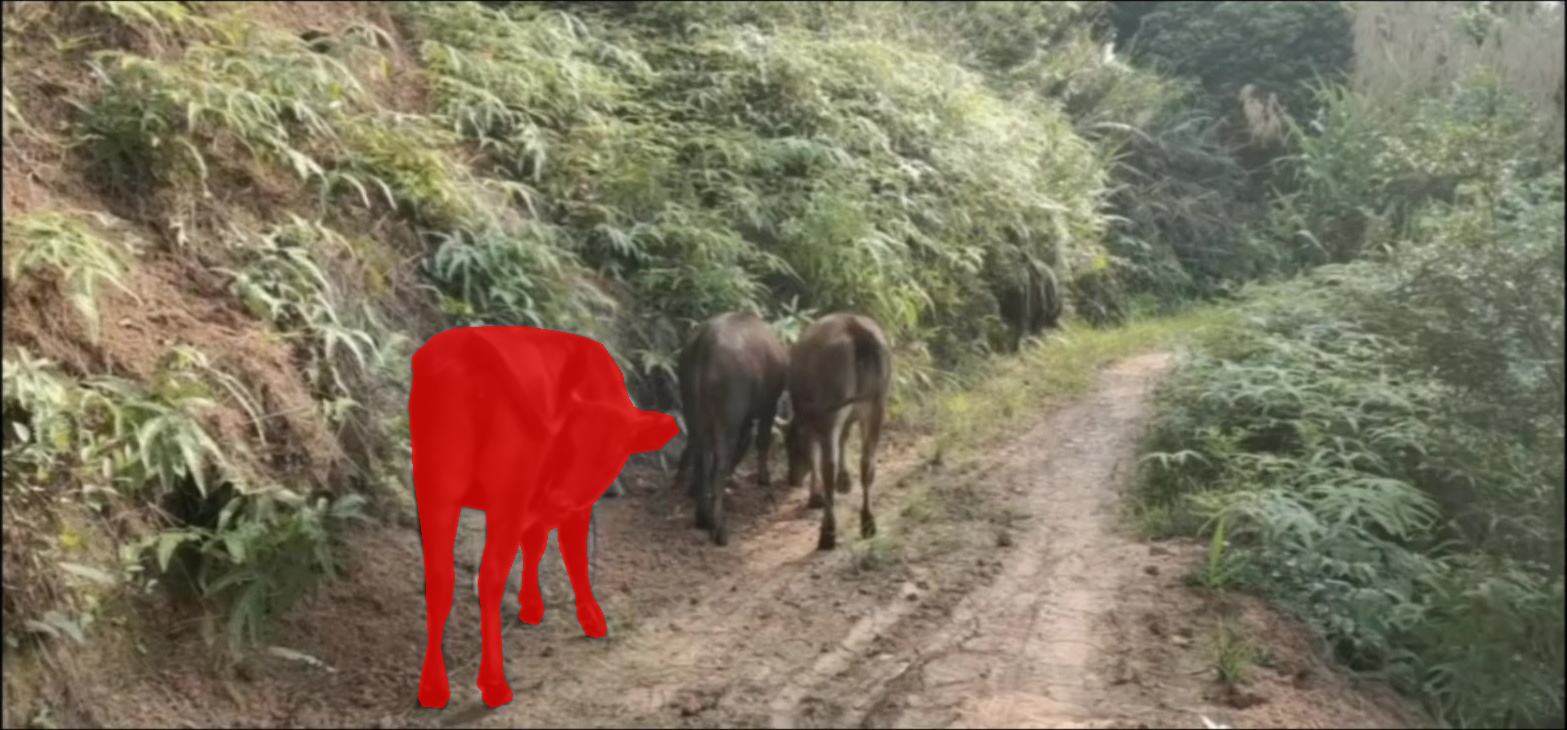}
\caption{}
\label{fig:keyframes-d}
\end{subfigure}%
\begin{subfigure}{0.235\textwidth}
\includegraphics[width=\textwidth]{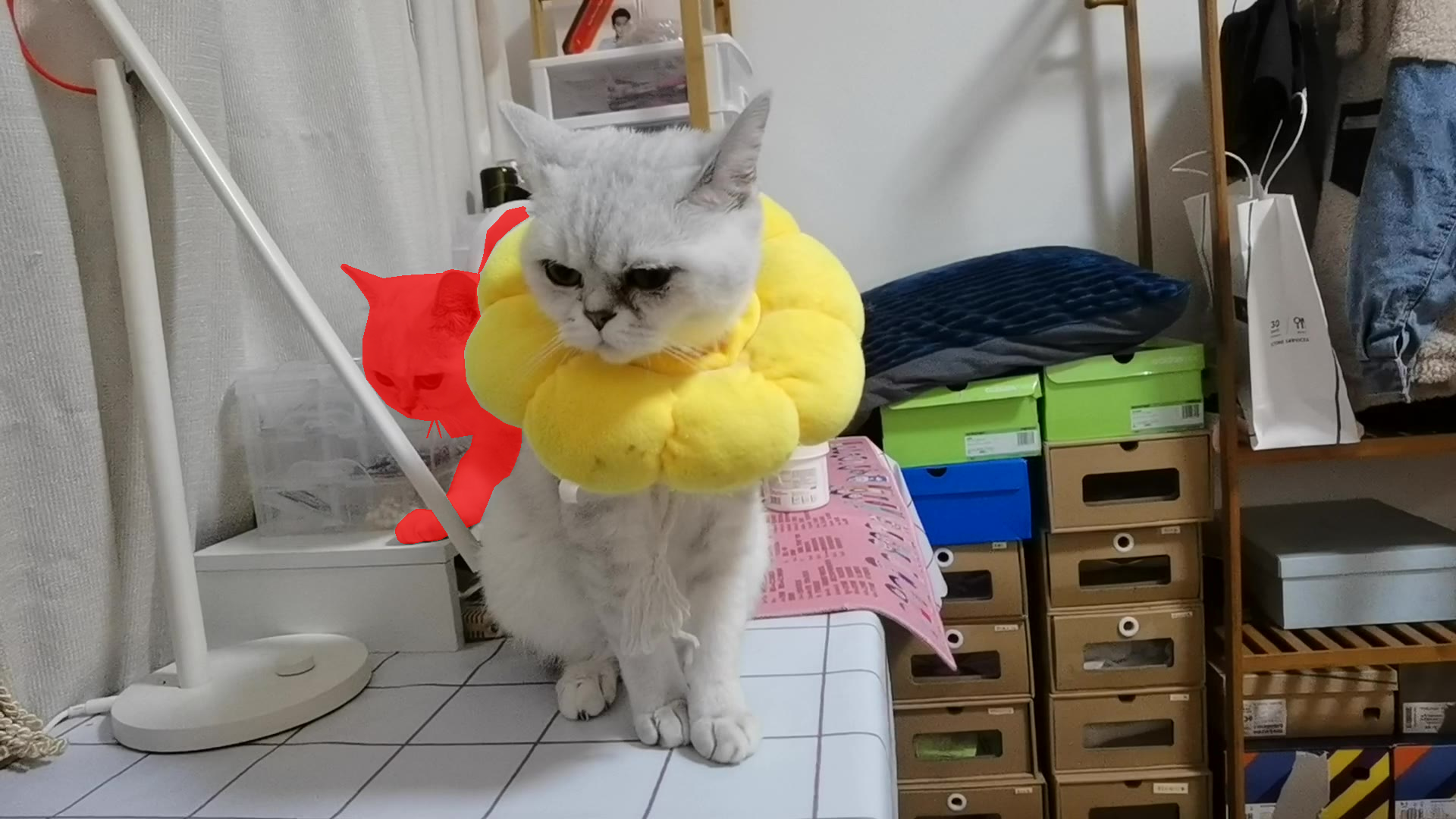}
\caption{}
\label{fig:keyframes-e}
\end{subfigure}
\vspace{-0.5em}
\caption{\textbf{Qualitative analysis of our proposed automatic keyframe selection} from five examples that show a single frame can be sufficient to convey the motion expression without any motion involved. It is mainly conveyed through the use of static cues such as the heading, object type, or position. Expressions of each example are as follows: (a) ``jump to the left then jump back'', (b) ``dog playing with monkey'', (c) ``puppy that overwhelms another puppy'', (d) ``cow shaking head and looking at us.'' (e) ``The little cat walking from behind to the front''. Groundtruth segmentation highlighted in red.} 
\vspace{-0.5em}
\label{fig:keyframes}
\end{figure*}

In this section, we summarize the shortcomings in the current video visual grounding benchmarks and video MLLMs. Visual grounding can be manifested in various tasks, including referring segmentation and grounded conversation generation, we focus on the former. Then, we describe our motion-centric probing and the respective adaptation to video MLLMs.

\subsection{Shortcomings in Video MLLMs and Visual Grounding Benchmarks}
 While there have been previous works focusing on establishing single-image baselines for video-language understanding on coarse-level tasks~\cite{buch2022revisiting}, (e.g., video question answering), we are the first to explore this within dense spatiotemporal grounding tasks. We focus on referring video segmentation and emphasize both standard and motion referring expressions~\cite{ding2023mevis}. We argue that the majority of referring expressions can be identified using strong single-image baselines that do not have an understanding of temporal information. While motion referring expressions may seemingly use motion, we show that such expressions can still be identified from one static frame. 

 Figure~\ref{fig:keyframes} shows five examples that can be sufficiently identified with static information without the use of temporal information, thus showing a weakness in the current evaluation benchmarks. Section~\ref{sec:results} confirms these findings quantitatively. In Figure~\ref{fig:keyframes-a},\ref{fig:keyframes-d},\ref{fig:keyframes-e}, we show examples that can be identified from one frame, where the direction, heading or the object's location in the referred expression can be deduced from one static frame. Moreover, in certain scenarios, stating the object category can be sufficient to ground the object without motion, e.g., Fig.~\ref{fig:keyframes-b} with only one dog in the scene. As such, we propose a motion-centric probing and evaluation to study the shortcomings in video MLLMs with respect to capturing the true motion. Our experiments confirm these limitations in prior benchmarks and methods. Additionally, we propose quantitative means to analyze the properties of referring expressions and videos that can drive a better evaluation for the use of motion beyond a single image.

\subsection{A Motion-Centric Probe and Dataset}
\label{sec:benchmark}

\begin{figure*}[t]
      \centering
      \includegraphics[width=0.9\textwidth]{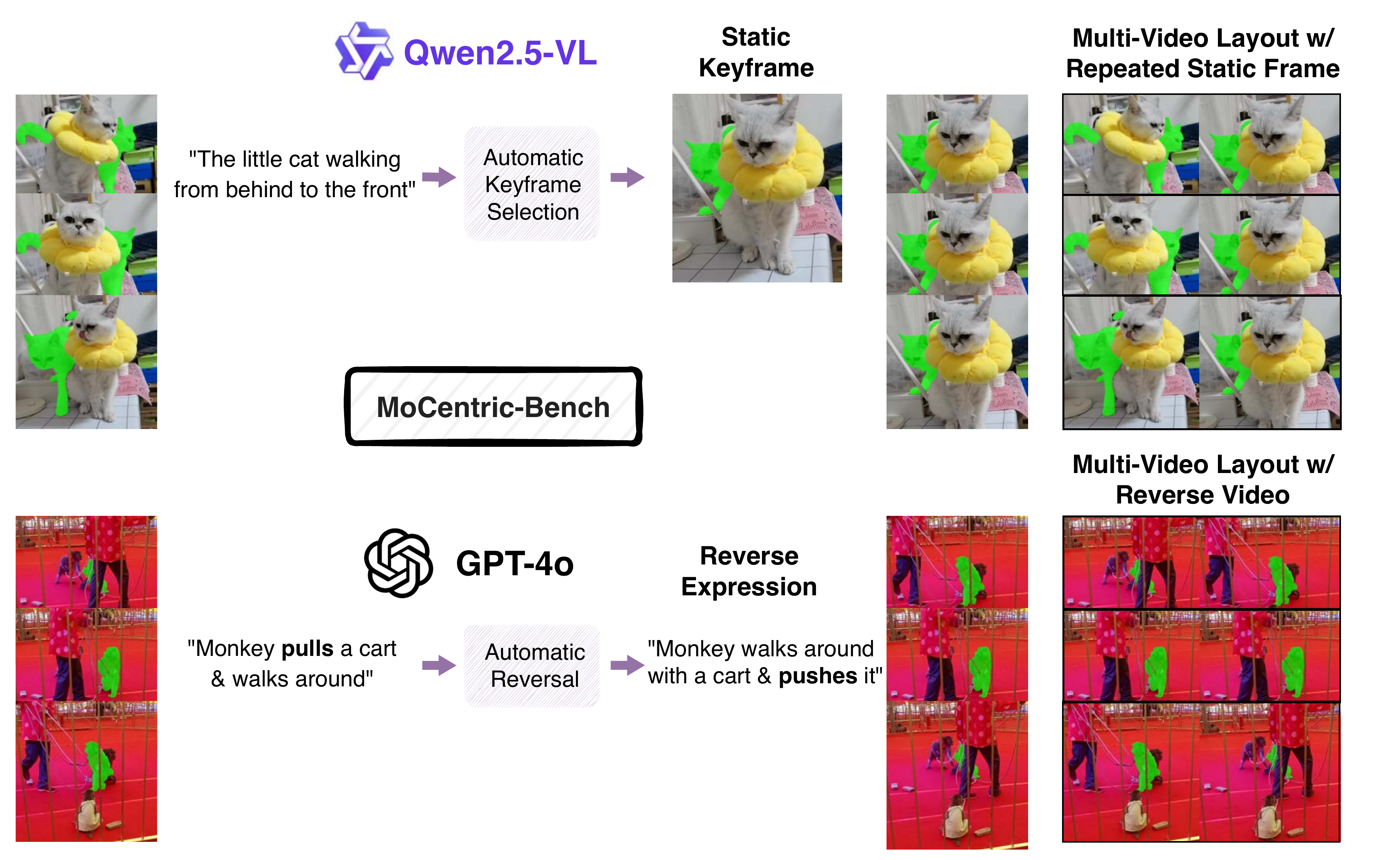}
  %     \vspace{-2.5em}
      \caption{\textbf{MoCentric-Bench.} The mechanism for creating our motion-centric benchmark. \textbf{Top:} Motion existence probe, $\mathcal{P}_1$. We rely on an automatic keyframe selection, using Qwen2.5-VL, to identify a static frame to convey the motion referring expression  (i.e., keyframe) and use it to deceive video-MLLMs as fake motion. The output modified video is the repeated keyframe and can be paired with the original video in various layouts to deceive their grounding mechanisms. We show that state-of-the-art models can easily be deceived with these static keyframes. \textbf{Bottom:} Motion ordering probe, $\mathcal{P}_2$. We use GPT-4o to perform an automatic motion referring expression reversal to match the respective reverse of the video. The output modified video and expression can similarly be paired with the original video to deceive video-MLLMs' grounding mechanisms in their ability to differentiate the original (forward) from the reverse of the video. $[.,.]$ is the horizontal concatenation operator. Ground-truth segmentation of the fake and true referred objects are highlighted in green.}
 \label{fig:detailed}
% \vspace{-1em}
\end{figure*}

 %\noindent \textbf{Motion-Centric Probing \& Synthetic Data.} 
 %\textcolor{red}{Introduce the four probing and explain why the tile specifically.}
Assume an input video with its respective motion expression and groundtruth masks, $S_i = (V_i, E_i, G_i)$, where $E_i$ is a referring expression, $V_i = (I^t)_{t=1}^{T}, G_i=(M^t)_{t=1}^{T}$ are a sequence of images, $I_t$ and their respective masks, $M_t$. We want to synthesize a modified sequence for the motion-centric probing, $\bar{S}_i = (\bar{V}_i, \bar{E}_i, \bar{G_i})$, with $\bar{V}_i = (\bar{I}^t)_{t=1}^{T}, \bar{G}_i=(\bar{M}^t)_{t=1}^{T}$. The  $j^{th}$ motion probe, $\mathcal{P}_j$, can result in, $\bar{S}^j_i = \mathcal{P}_j(S_i)$. We describe two motion probes, yet we foresee our method can be extended beyond these: (i)  \textit{Motion Existence}, $\mathcal{P}_1$: is to inspect video-MLLMs' ability to identify true motion across the video vs. fake motion from one repeated static single frame, and (ii) \textit{Motion Ordering}, $\mathcal{P}_2$: probes video-MLLMs for their ability to ground the expression and differentiate it from its reverse and vice versa. Figure~\ref{fig:detailed} describes the two motion probes and how we construct their modified sets. 
 %The motion existence probe results in two simple ways to inspect video-MLLMs. The first is using the video with a static single frame repeated without any motion as input. This frame should be roughly the best in conveying the motion referring expression; we refer to it as the keyframe. The second is using a multi-video layout that includes both the original video and the video with that repeated static keyframe, in various layouts, as described earlier. The motion order results in another two simple probes. The first uses the reverse of the video along with the reverse of the motion referring expression, while the other uses a similar multi-video layout with both the original video and the reverse one while querying with the reverse motion expression. 

In the motion existence probe, an automatic selection of the keyframe is used. The goal is to identify a single static frame that can sufficiently mislead video MLLMs, to capture the motion expression without temporal information. Towards that, we use a multi-modal large language model that has the capability of coarse video grounding, i.e., identifying the events' segments in seconds guided by language descriptions without spatial grounding. We prompt the MLLM to identify the static frame using the following: \textit{``Given the query: \texttt{<EXP>}, when does the described content occur in the video? Output the first and last seconds for this action in JSON format.''}, where \texttt{<EXP>} is the motion referring expression. The output is further processed to identify the temporal window, where its middle frame is labelled as the keyframe capturing this motion as a proxy,  as shown in Figure~\ref{fig:detailed} (top). As such,  the modified sequence, $\bar{S}^1_i$,  has the repeated static keyframe and corresponding empty masks, while the same motion expression is used. Figure~\ref{fig:keyframes} shows five example keyframes automatically selected from their motion referring expression. 
%We use the retrieved keyframes to create a video containing both the static keyframe repeated, in addition to the original video,) 

In the motion ordering probe, the crucial step is to reverse the motion referring expression to match the reversal of the video. Towards that, we prompt a large language model using the following: \textit{``Taken an input motion referring expression as \texttt{<EXP>} corresponding to an original video, can you convert it to the motion expression describing what will occur in the reverse version of the same video? Output the new expression only.''}. This is followed by manual inspection of the reverse motion referring expressions with their respective reverse videos, to discard examples that can not be differentiated from the original motion using the expression. Then,  the modified sequence, $\bar{S}^2_i$, has the reverse video, the reversed groundtruth masks and the reversed expression. We further change the masks, $G_i$, for the original sequence to empty masks, since the referring expression is only inquiring about the reverse motion. Figure~\ref{fig:detailed} (bottom) shows an example of the automatic reversal for the motion expression from \textit{``pulling''} to \textit{``pushing''}. 

Each original sequence, $S_i = (V_i, E_i, G_i)$, and its modified variant, $\bar{S}^j_i = (\bar{V}^j_i, \bar{E}^j_i, \bar{G}^j_i)$, is further processed to generate the final sequence based on a specific layout. We leave out the notation for the specific probe, $.^j$, for simplicity when describing the layouts. We detail three layouts we use that challenge visual grounding models, yet our synthetic data can be extensible beyond these three: (i)  $l_1$ follows \eqref{eq:1a}, (ii)  $l_2$ follows \eqref{eq:1b} and (iii)  $l_3$ follows \eqref{eq:1c},

\begin{subequations}
\begin{equation}
 \bar{S}_i = (\bar{V}_i, \bar{E}_i, \bar{G}_i),
 \label{eq:1a}
 \end{equation}
 \begin{equation}
  \bar{S}_i =( ([I^t_i, \bar{I}^t_i])_{t=1}^T, \bar{E}_i, ([M^t_i, \bar{M}^t_i])_{t=1}^T),
  \label{eq:1b}
 \end{equation}
 \begin{equation}
 \bar{S}_i = ( ([\bar{I}^t_i, I^t_i])_{t=1}^T, \bar{E}_i, ([\bar{M}^t_i, M^t_i])_{t=1}^T), 
 \label{eq:1c}
\end{equation}
\end{subequations}
 
\noindent where, $[.,.]$ is the horizontal concatenation operator. The latter two layouts, $l_2, l_3$, are what we consider a multi-video layout. This results in four ways to probe the model: (i) repeated single frame video following  $\mathcal{P}_1$ and $l_1$, (ii) the reverse of the original video following $\mathcal{P}_2$ and $l_1$, (iii) multi-video layout with the repeated single frame following $\mathcal{P}_1$ and $l_2, l_3$ and (iv) multi-video layout with the reverse video following  $\mathcal{P}_2$ and $l_2, l_3$. These can help inspect the video-MLLMs' true ability in motion and video understanding within a visual grounding framework. 

%Both $l_2$ and $l_3$ layouts can introduce failures in video MLLMs that are due to spatial biases beyond the failures in capturing local motion patterns. We ensure our evaluation is focused on the failures in the local motion patterns in four ways. The two motion probes have different regions in the video to focus on. Assuming the second layout, $l_2$, with $\mathcal{P}_1$, the segmented region is focused on the right side with $V_i, M_i$, while $\mathcal{P}_2$ is focused on the other side, $\bar{V}_i, \bar{M}_i$. As such, when evaluating both, we avoid being skewed to spatial biases for a certain region in the input video. Our strong single-image baselines are evaluated following both $l_2$ and $l_3$ in each motion probe,  and the predictions from both layouts are combined for their evaluation. Similarly, we provide an evaluation for Sa2VA variants that averages the predictions from both layouts in the supplemental. Moreover, we report the false positive rate on the deceiving region of the video in the supplement. Finally, we report evaluation on the \textit{Single frame} variant to confirm that state-of-the-art video MLLMs exhibit failures in identifying local motion patterns guided by language expressions, without any spatial biases. Further details are provided in the supplementary material. 

\begin{table}[t]
\caption{\textbf{Comparison of our mocentric dataset with visual grounding benchmarks.} It confirms that our benchmark maintains close or on-par statistics while introducing the advantage of evaluating MLLMs' ability to capture local motion. Note that locality here refers to the temporal dimension, where a single frame or non-consecutive keyframes are insufficient to convey the motion expression without a genuine understanding of the underlying motion between them.}
\resizebox{\textwidth}{!}{
\begin{tabular}{l|c|c|c|c|c}
\toprule
 \textbf{Datasets \& Benchmarks} &  \textbf{Venue} &  \textbf{Videos} & \textbf{Expressions}  &   \textbf{Annot. Type} &  \textbf{Motion Probe (Local)} \\ \midrule
\multicolumn{6}{l}{\textit{Training Datasets w/ or w/o Evaluation.}} \\ \midrule
\rowcolor{gray!50}RefDAVIS~\cite{khoreva2018video}                  & ACCV 2018  & 90    & 1,544    &  Mask                      & \xmark                               \\
\rowcolor{gray!50} RefYTVOS~\cite{seo2020urvos}                    & ECCV 2020 & 3,978               & 15,009 &  Mask     & \xmark                                \\
\rowcolor{gray!50}MeVIS~\cite{ding2023mevis}          & ICCV 2023  & 2,006 & 28,570   & Mask                        &  \xmark                              \\ 
\rowcolor{gray!50}ReVOS~\cite{yan2024visa} & ECCV 2024  & 1,042 & \textbf{35,074}  & Mask                        &  \xmark 		\\
\rowcolor{gray!50}MeVIS v2~\cite{ding2025mevis}          & TPAMI 2025  & 2,006 & 33,072    & Mask                        &  \xmark                              \\ 
Molmo2VideoTrack - Box~\cite{clark2026molmo2}    & CVPR 2026 & 4,059 (clips)  & 19,653 &  Box          &  \xmark                               \\ 
\rowcolor{gray!50}Molmo2VideoTrack - Mask~\cite{clark2026molmo2}    &  CVPR 2026 & 2,565 (clips)  & 10,051  & Mask                           &  \xmark                               \\  \midrule
\multicolumn{6}{l}{\textit{Evaluation Only Benchmarks.}} \\ \midrule
\rowcolor{gray!50}ReasonVOS~\cite{bai2024one}                  & NeurIPS 2024 & 105  & 458   &  Mask                       &  \xmark                               \\ 
Molmo2Track - Box~\cite{clark2026molmo2}   & CVPR 2026 & 1,358 (clips) & 3,067 &  Box                           & \xmark                                \\
\rowcolor{gray!50}Molmo2Track - Mask~\cite{clark2026molmo2}   & CVPR 2026 & 28 (clips)  & 80   &  Mask                           &  \xmark                               \\  \midrule \midrule
\rowcolor{gray!50} MoCentric-Mask - \textit{train \& eval.}              & Ours (NeurIPS 2026)  & \textbf{3,658 (modified videos)}  &  27,654  &  Mask                    & \qquad  \qquad \quad  \textbf{\cmark}  \multirow{2}{*}{ 
\resizebox{0.1\textwidth}{!}{
\begin{tikzpicture}[baseline=(X.base)]
    % Adding align=center unlocks the use of \\ for newlines
    \node[draw=red, fill=white, rounded corners, inner sep=6pt, thick, align=center] (X) at (0, 0.5) {
        Existence \& \\ Ordering
    };
    \useasboundingbox (0,0); 
  \end{tikzpicture}%   
 } }\\
 MoCentric-Box  - \textit{eval. only}             & Ours (NeurIPS 2026)  & 188 (modified videos)   &  332  & Box                    &   \textbf{\cmark}       \\\bottomrule                   
\end{tabular}}
\vspace{-1em}
\label{tab:mocentric_data_soa_comparison}
\end{table}

\textbf{Dataset quality and statistics.} We use the aforementioned probing to construct synthetic training data, relying on the training subset of publicly available referring video segmentation datasets and a motion-centric benchmark, MoCentric-Bench, using their validation/test subsets. Our proposed probing yields synthetic data comprising 3,846 videos and 27,986 expressions. To confirm the automatic data annotations used in our probing, we perform a verification phase with two annotators who rate our benchmark. Each annotator is requested to score the referring expression and its relation to and description of the video, with a focus on the static key-frame or the reverse video. Each video-expression pair is rated with a score of 1-3, with 1 as the best and 3 as the worst. Our verification resulted in 87.8\% of the video-expression pairs corresponding to a score of 1, respectively, where the majority of the expressions and keyframes align with the intended use. Score 3, which is the lowest quality, is only at 4.1\%. We also report inter-annotator agreement of 83.7\% with absolute score agreement. Finally, we report the Cohen’s Kappa score at 0.43, which is acceptable for a subjective 3-point scale ranking, as shown in previous works~\cite{ghosh2023geneval}. 

MoCentric-Bench evaluation occurs on $l_2$ layout, yet $l_3$ is used during inference as well. Both \( l_2 \) and \( l_3 \) layouts can introduce failures in video MLLMs that stem from \textit{spatial biases}, beyond those related to capturing \textit{motion patterns}. To ensure that our evaluation isolates failures specific to motion understanding, we take three measures. First, we include results of spatial bias probing that uses a blank single colour image in a multi-video layout with the original video, and compare the performance degradation from our motion probes vs. the spatial biases. The results show that state-of-the-art video MLLMs exhibit failures in recognizing local motion patterns guided by language expressions, removing any factors from spatial biases. Second, our strong single-image baselines are evaluated under both \( l_2 \) and \( l_3 \) layouts for each motion probe, and their predictions are combined for the final evaluation on $l_2$ layout. Similarly, for state-of-the-art models, we report results that average predictions across both layouts. Third, we report the false positive rate on the deceiving region of the video. Refer to Appendix~\ref{sec:more_results} for these results and discussion.

Table~\ref{tab:mocentric_data_soa_comparison} compares different benchmarks in the visual grounding literature, confirming that our benchmark maintained on-par statistics and scale with respect to the ones providing mask-level annotations, while introducing the advantage of evaluating MLLMs' ability to capture local motion. In previous benchmarks, a single keyframe or a few non-consecutive ones are sufficient to visually ground the motion expression, without necessarily conveying the proper local motion. On the other hand, our benchmark pushes MLLMs to differentiate locally, forward from reverse motion and true from fake/apparent motion conveyed in a single static frame. Our probe with the deceiving region can confuse such MLLMs to ground these motion expressions, even when the local motion does not match it. Hence, our benchmark is better at evaluating their ability to capture the motion expression in relation to its language description, even locally within a few keyframes. 

\textbf{Discussion of alternatives.} One potential alternative that was explored in action recognition is to use synthetic random dot cinematographs~\cite{ilic2022appearance}. However, we found these to introduce severe distribution shifts that would require re-training or finetunng such video MLLMs. The aforementioned training can introduce its own pitfalls in such general-purpose models, especially if not implemented natively, and is beyond the scope of evaluating MLLMs ability to capture motion. Our probing avoids this distribution shift, relying on natural videos, yet synthesizing such multi-video layouts. Multimodal video understanding training data, curated from the internet, have multi-video layouts naturally occurring (e.g., InternVid~\cite{wang2024internvid}),  since split-screens and PIP are used in news, social media videos, and others. Another alternative is relying on videos with camouflaged objects moving, yet previous literature~\cite{kowal2022deeper, karim2023med++} has shown that it still requires appearance understanding. It does not fully decouple appearance from motion without using interpretability mechanisms that are outside the scope of this benchmark.

%Concurrent work from Molmo2Track has shown a larger evaluation benchmark, yet it did not resolve the issue tackled in this work regarding assessing motion and language interaction. I need to add Molmo2Track as another dataset. Table~\ref{tab:mocentric_bench_stats} shows statistics of the aforementioned constructed datasets for both the evaluation and the training.

%%%%%%%%%%%%%%%%%%%%%%% Count number of expressions not video-expression pairs to match the table above so its well explained

% \noindent \textbf{Motion-Centric Dataset Curation.} 
 %\textcolor{red}{explain how to curate the GUI grounding data as synthetic with different motions patterns, similarly for Autonomous Driving and Robot Manipulation. }
 
% \begin{table}[]
%\begin{tabular}{l| c| c| c| c| p{20mm}}
%\hline
%Benchmark & Pub-Year & \# Videos & \# VQA Pairs & Motion Ex. & Setting  \\ \hline
%RefDAVIS$_{16}$~\cite{} & ACCV 2018 &  50 &  & - & Generic \\ 
%RefDAVIS$_{17}$~\cite{} & ACCV 2018 &  90 &  & - & Generic\\ 
%Refer-YouTube-VOS~\cite{} & ECCV 2020 & & & - &  Generic \\ 
%MeVIS$_{valid_u}$~\cite{} & ICCV 2023 & 50 & 793 &\checkmark & Generic  \\ 
%MeVIS$_{valid}$~\cite{} & ICCV 2023 & 140 & &\checkmark & Generic  \\ 
%VPoS Bench~\cite{} & Arxiv 2025 & & 100 &  46 &AutoDrive, Robot Manip., Ego, GUI \\ \hline
%MoCentric-Bench (ours) & 2025 & 500 &   & \checkmark & Generic, AutoDrive, Robot Manip., GUI\\  \hline
%\end{tabular}
%\caption{Evaluation Benchmarks.}
%\end{table}

\subsection{Baselines and Motion-Centric Adaptation}
\textbf{Strong baselines.} We propose strong single-image baselines for referring video segmentation using powerful MLLMs that can visually ground objects on the region level. Models such as Qwen2.5-VL~\cite{bai2025qwen2} and InternVL3~\cite{zhu2025internvl3} have emerged that show strong capabilities in visually grounding objects and outputting their corresponding bounding boxes. We build two baselines that are biased to the static information conveyed from a single image. The first baseline relies on the MLLM output with the first frame in the video to initialize SAM 2.0~\cite{ravi2024sam} and generate the output segmentation for the corresponding referring expression (Baseline).  The second baseline follows a similar procedure, but instead of using the first frame, it uses the keyframe in the input video to initialize SAM 2.0 (Baseline$\dagger$). The keyframe is retrieved following the previous method in the motion-centric probing. %Further implementation details are described in Appendix~\ref{sec:imp_details}.

\textbf{Motion-centric adaptation.} We provide a simple mechanism to adapt pixel-level multi-modal large language models to improve their abilities in identifying local motion. We hypothesize that the reason behind the models' inability to identify true vs. fake motion or the forward vs. the reverse motion is tied to a weak ability to capture motion itself, in addition to its interaction with language. Thus, we perform a motion-centric low-rank (LoRA) adaptation~\cite{hu2022lora} of the vision encoder and connector using supervised finetuning data that is built with our motion-centric probing. During the LoRA fine-tuning, the weights of the LLM and the segmentation decoder, if present, are fixed. Additional implementation and evaluation details are provided in Appendix~\ref{sec:imp_details}.%Refer to Appendix~\ref{sec:imp_details} for details. 

\section{Experimental Results} % 3 page
\label{sec:results}
\subsection{Experimental Setup}
\noindent \textbf{Datasets and metrics.} We use referring video segmentation datasets in our supervised finetuning, including RefDAVIS17~\cite{pont20172017}, MeVIS~\cite{ding2023mevis}, referring YouTubeVOS~\cite{seo2020urvos}, the reasoning VOS dataset~\cite{yan2024visa} and our motion-centric variants from MeVIS \textit{training}. For the evaluation, we evaluate on the \textit{validation} split of RefDAVIS17 and for MeVIS, we use two splits: the \textit{val$^u$} subset, which includes 50 videos with their ground-truth and the \textit{val} subset, which includes 140 videos that can be evaluated using the evaluation server~\footnote{\url{https://codalab.lisn.upsaclay.fr/competitions/15094}}. Finally, we evaluate on our constructed motion-centric variants of MeVIS \textit{val$^u$}, as detailed in Sec.~\ref{sec:benchmark}, which are: (i) \textit{Single frame} following $l_1$ layout and $\mathcal{P}_1$ probe, (ii) \textit{val$^u$ \& Single frame} following $l_2$ and $\mathcal{P}_1$, (iii) \textit{Reverse} following $l_1$ and $\mathcal{P}_2$, and (iv) \textit{val$^u$ \& Reverse} following $l_2$ and $\mathcal{P}_2$. The \textit{Reverse} variants of MoCentric-Bench include 28 videos and 152 segmentation-motion expression pairs. While the \textit{Single frame} variants have 50 videos and 793 segmentation and expression pairs. We use standard metrics for the region similarity, $\mathcal{J}$, the contour accuracy, $\mathcal{F}$, and their average, $\mathcal{J\&F}$.
% It seems it benefits from random L & R Static but no Reverse. Seems the reverse is giving it hard time bec. no filtration during training on examples that are
% sufficiently differentiated between forward vs. reverse video unambiguously.

\noindent \textbf{Compared methods.} We compare our strong baselines, Baseline and Baseline$\dagger$, to state-of-the-art referring video segmentation methods, including the prior ones that relied on masked modelling from the RoBERTa model, such as ReferFormer~\cite{wu2022language} and LMPM~\cite{ding2023mevis}. Additionally, we compare against recent ones that rely on the power of large language models with autoregressive modeling in LISA~\cite{lai2024lisa}, TrackGPT~\cite{zhu2023tracking}, VISA~\cite{yan2024visa}, VideoGLAMM~\cite{munasinghe2024videoglamm}, RGA~\cite{wang2025object}, Sa2VA~\cite{yuan2025sa2va} and Molmo 2.0~\cite{clark2026molmo2}. For the motion-centric evaluation, we focus specifically on the best models in each category, which have their codes and weights publicly available (i.e., LMPM, VideoGLAMM, RGA, Sa2VA and Molmo 2.0), in addition to our strong baselines and our motion-centric adapted ones (i.e, Sa2VA$\star$, Molmo 2.0$\star$).

%ReferFormer~\cite{wu2022language}, LMPM~\cite{ding2023mevis} , LISA~\cite{lai2024lisa}, TrackGPT~\cite{zhu2023tracking} , VISA~\cite{yan2024visa} , VidGLAMM~\cite{munasinghe2024videoglamm}, Sa2VA~\cite{yuan2025sa2va} , LLaVA-7B~\cite{liu2023visual}, ChUniVi-7B~\cite{jin2024chat}, VE+Phi3-mini-3.8B~\cite{abdin2024phi3technicalreporthighly} , InternVL2-8B~\cite{chen2024expanding}, Qwen2.5-VL-7B~\cite{bai2025qwen2} 
%', (ii) second column is ``black one that turns and goes left''.} 

\begin{table*}[t]
\centering
\caption{\textbf{MoCentric-Bench} results based on the synthesized multi-video layout variants of MeVIS dataset \textit{val$^u$} subset, following probe $\mathcal{P}_1$ and $l_2$ layout (i.e., \textit{val$^u$ \& Single frame}), following probe $\mathcal{P}_2$ and $l_1$ (i.e., \textit{Reverse}) and following probe $\mathcal{P}_2$ and $l_2$  (i.e., \textit{val$^u$ \& Reverse}). $\dagger$: indicates the use of the automatic keyframe selection. $\star$: the motion-centric adaptation. It clearly shows a drop in performance for all the models, including our baselines from \textit{val$^u$} to \textit{val$^u$ \& Single frame}, similarly between the \textit{Reverse} and  \textit{val$^u$ \& Reverse}. It shows that the state-of-the-art methods and our single-image baselines can not differentiate true from fake motion and do not necessarily understand the ordering of motion. Best results are bolded.}
\vspace{-0.5em}
\resizebox{\textwidth}{!}{
\begin{tabular}{l|c|ccc|ccc|ccc|ccc}
\toprule
Method & Venue & \multicolumn{3}{c|}{ \textit{val$^u$} }&  \multicolumn{3}{c|}{ \textit{val$^u$ \& Single frame}} & \multicolumn{3}{c|}{\textit{Reverse}} & \multicolumn{3}{c}{\textit{val$^u$ \& Reverse} }  \\
 &  & $\mathcal{J}$ & $\mathcal{F}$ & $\mathcal{J\&F}$ & $\mathcal{J}$  & $\mathcal{F}$ & $\mathcal{J\&F}$ & $\mathcal{J}$ & $\mathcal{F}$ & $\mathcal{J\&F}$ & $\mathcal{J}$ & $\mathcal{F}$ & $\mathcal{J\&F}$\\ \midrule
 %\multicolumn{14}{l}{\textbf{Closed Source MLLMs}} \\ \midrule
GPT-5 &  - & 30.0 & 36.9 & 33.5 & 17.8 & 27.8 & 22.8 & 33.9 & 39.8 & 36.8 & 17.5 & 26.3 & 21.9 \\ \midrule
%Gemini-3 Flash &  - & - & -  & 51.1 &  &  &  &  & & &  & & \\ \midrule
 %\multicolumn{14}{l}{\textbf{Open Source Video/Image MLLMs}} \\ \midrule
LMPM & ICCV 2023 & 34.2 & 40.2 & 37.2 & 18.2 & 26.9 & 22.6 & 31.8 & 38.1 & 35.0 & 18.0 & 25.4 & 21.7\\
VidGLAMM & CVPR 2025 & 43.6 & 52.7 & 48.2  & 18.1 & 25.1 & 21.6 & 49.4 & 58.4 & 53.9 & 28.7 & 39.3 & 34.0 \\  
RGA & ICCV 2025 & 53.2 & 59.1 & 56.1 & 23.9 & 32.5 & 28.2 & 57.3 & 63.3 & 60.3 & 32.0 & 42.1 & 37.0 \\  % Check Reverse results probably wrong % 59.7 cs 56.1
%Qwen3-VL (video) + SAM 2.0 & Arxiv 2025 & 32.7& 44.7 & 38.7 & &  &  &   &  &  &  &  &  \\ 
Sa2VA & TPAMI 2026 & - & - & 58.9  & 23.5 & 33.5 & 28.5 &  55.9 & 66.4 & 61.1 & 31.0 & 42.9 & 37.0 \\
%Molmo2 & CVPR 2026 & 61.0 & 70.0 & 70.8/65.4 & 32.9 & 47.0 & 40.0 & 59.1 & 66.7 & 62.9 & 25.3 & 42.8 & 34.1 \\  \midrule%\midrule OLD Results
Molmo 2.0 + SAM 2.0 & CVPR 2026 & - & - & 70.0 & 33.1 & 47.8 & 40.5 & 59.3 & 66.9 & 63.1 & 24.5 & 41.7 & 33.1\\  \midrule
%%%%%%%%%% The current active experiments%%%%%%%%%%%%%%%%%%%%%%%%%%%%%%%%%%%%%%%%%%%%%%%%%%%%%%%%%%%%%%%%%%%%%%%%%%%%%%%%%%
%Molmo2 run2 & CVPR 2026 & - & - & 70.0 &  &  &  &  &  &  &  &  & 31.3\\  
%Molmo2 - LoRA Base & CVPR 2026 & - & - &  & & &  &  &  &   & &  & 32.9 \\ 
%%%%%%%%%% Other experiments%%%%%%%%%%%%%%%%%%%%%%%%%%%%%%%%%%%%%%%%%%%%%%%%%%%%%%%%%%%%%%%%%%%%%%%%%%%%%%%%%%%%%%%%%
%Molmo2 - LoRA MoCentric only  & CVPR 2026 & - & - &  & & &  &  &  &  & &  & \\ 
%Molmo2 - LoRA MeVIS variants & CVPR 2026 & - & - &  & & &  &  &  &  & &  & \\ 
%Molmo2 - LoRA Base all & CVPR 2026 & - & - &  & & &  &  &  &  & &  & \\ 
%Molmo2 - LoRA MoCentric all & CVPR 2026 & - & - &  & & &  &  &  &  & &  & \\ 
%Molmo-Motion & Arxiv 2026 & - & - &  & & &  &  &  &   & &  & \\ \midrule % without scaling down
%ReferMo & CVPR 2026 &  &  &  &  &  &  &  &  &  &  &  & \\  % low performance but might be interesting from the part it is taking input motion frames similar to OF
%VIRST & CVPR 2026 &  &  &  &  &  &  &  &  &  &  &  & \\  % performance on-par to Molmo2
%%%%%%%%%% END%%%%%%%%%%%%%%%%%%%%%%%%%%%%%%%%%%%%%%%%%%%%%%%%%%%%%%%%%%%%%%%%%%%%%%%%%%%%%%%%%%%%%%%%%%%%%%%%%
%MLLM+S2 & 47.9 & 56.2 & 52.0  & 17.0  & 25.5 & 21.3 & 44.0 & 53.0 & 48.5  &24.2  & 32.4  & 28.3 \\   %Old result w/ wrong flip  
Baseline & ours (NeurIPS 2026) & 47.9 & 56.2 & 52.0  &  21.4 & 31.6 & 26.5 & 44.0 & 53.0 & 48.5  & 21.4 & 30.2  & 25.8 \\
%MLLM+S2$\dagger$ & 54.0 & 60.9 & 57.4  & 19.5 & 26.7 & 23.1 & 49.5 & 57.3 & 53.4  & 31.1 & 39.5 & 35.3 \\ %Old result w/ wrong flip
Baseline$\dagger$ & ours (NeurIPS 2026) & 54.0 & 60.9 & 57.4  & 23.0 & 33.2 & 28.1 & 49.5 & 57.3 & 53.4  & 26.1 & 36.1 & 31.1 \\  
%Qwen3-VL (single-image) + SAM 2.0 & ours (2025) & 12.4 & 17.5 & 14.9 & &  &  &   &  &  &  &  &  \\ 
%Qwen3-VL (single-image) + SAM 2.0$\dagger$ & ours (2025)& 14.5 & 19.4 & 16.9 & &  &  &   &  &  &  &  &  \\ 
%Sa2VA$\star$ - Run1& & 55.5 & 64.7 & 60.1   & 25.0 & 35.3  & 30.1 & 61.3 & 70.4 & 65.9 & 34.6 & 48.5 & 41.6 \\ % Run1
%Sa2VA$\star$  - Run2 & & 56.3 & 65.3 & 60.8 & 26.6 & 37.3  & 32.0 & 59.2 & 69.4 & 64.3 &  35.5 & 49.0 & 42.3  \\ \midrule % Run2
%Sa2VA$\star$ My Base&- &- & - & - &  - & 31.8 & - & - & - & - &- & 34.8  \\ \bottomrule % My base an improved version
%Sa2VA$\star$ Single+Reverse &- &- & - & - &  - & 30.8  & - & - & - & - &- & 42.1  \\ \bottomrule % Using Single and Rev not Single only
Sa2VA$\star$ & ours (NeurIPS 2026) & 55.9 & 65.0 & 60.5  & 25.8 &  36.3 & 31.1 & 60.3 & 69.9 & 65.1 &  \textbf{35.1} & \textbf{48.8} & \textbf{42.0}  \\  
%Sa2VA$\star$ & Release Run & 57.0&65.2 & 61.1  & 26.5 & 37.6 & 32.1 &61.7 & 70.4 & 66.0 &  32.4 & 45.7 & 39.0  \\  
Molmo 2.0$\star$ + SAM 2.0 & ours (NeurIPS 2026)  & \textbf{66.4} & \textbf{73.8} & \textbf{70.1} & \textbf{42.5} & \textbf{54.7} & \textbf{48.6} & \textbf{62.4} & \textbf{70.5} & \textbf{66.6}  & 30.7 & 45.4 & 38.1 \\ \bottomrule
\end{tabular}}
\label{table:mocentric}
\vspace{-1em}
\end{table*}

\subsection{MoCentric-Bench Results}
In this section, we start with the motion-centric evaluation, where we show that both the state-of-the-art methods and our strong baselines still fall short in identifying true vs. fake motion and in understanding the ordering of motion. Table~\ref{table:mocentric} shows the evaluation on the \textit{val$^u$} set of MeVIS and the corresponding motion-centric variants. Across all the methods, there is a clear degradation of around half the original performance on the standard videos (i.e., \textit{val$^u$} or \textit{Reverse}) vs. the multi-video layout ones that are quite challenging for visual grounding (i.e., \textit{val$^u$ \& Single frame} or \textit{val$^u$ \& Reverse}). It highlights the major shortcomings in state-of-the-art visual grounding video MLLMs, where most of the methods do not have a proper understanding of motion and are mostly biased toward information conveyed from a single frame. Hence, we show with sufficient evidence the failures of video MLLMs in capturing local motion and being deceived by our probing.
%Supplementary Table S3 confirms this further, evaluating state-of-the-art models with the repeated \textit{Single frame} only. 

Our motion-centric adapted variants, Sa2VA$\star$ and Molmo 2.0$\star$, provide state-of-the-art performance across the four settings. When looking at the gap between the multi-video layout and its respective original evaluation, e.g., Sa2VA vs. Sa2VA$\star$, we notice an improvement of 1- 8\%, where our motion-centric adaptation reduces this gap. In Appendix~\ref{sec:more_results}, Table~\ref{table:fpr_mocentric} we report the false positive rate in the deceiving region across two settings to confirm the benefit from our simple motion-centric adaptation. However, there remains significant potential for improvement through developing MLLMs that are both vision and motion centric in their core, beyond simple adaptation mechanisms. Additional results ablating the frame selection are provided in Appendix~\ref{sec:more_results}.

\begin{figure*}[t]
\centering
\resizebox{0.8\textwidth}{!}{
\begin{subfigure}{0.31\textwidth}
\includegraphics[width=\textwidth]{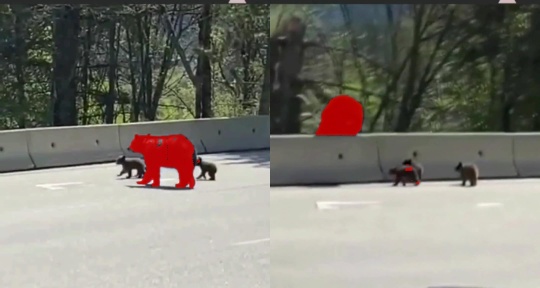}
\end{subfigure}%
\hspace{0.2em}%
\begin{subfigure}{0.31\textwidth}
\includegraphics[width=\textwidth]{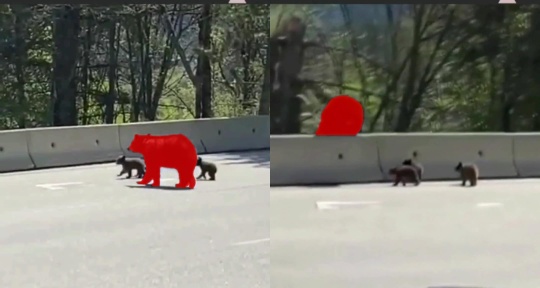}
\end{subfigure}%
\hspace{0.2em}%
\begin{subfigure}{0.31\textwidth}
\includegraphics[width=\textwidth]{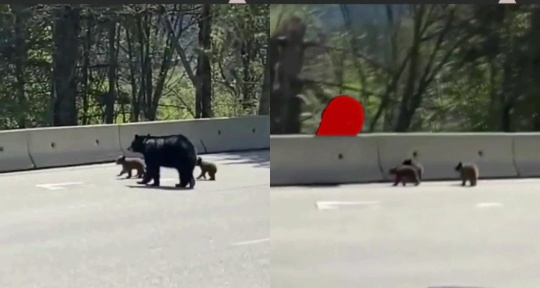}
\end{subfigure}}

\resizebox{0.8\textwidth}{!}{
\begin{subfigure}{0.31\textwidth}
\includegraphics[width=\textwidth]{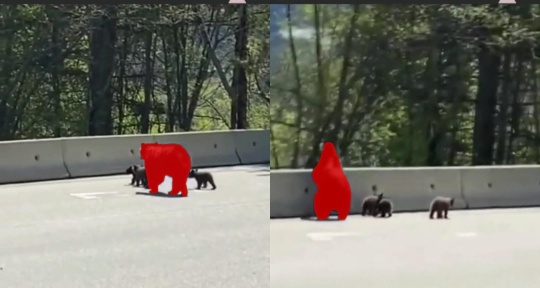}
\end{subfigure}%
\hspace{0.2em}%
\begin{subfigure}{0.31\textwidth}
\includegraphics[width=\textwidth]{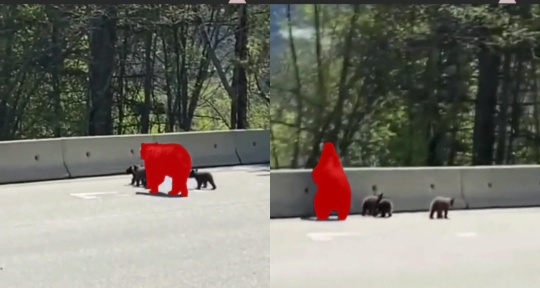}
\end{subfigure}%
\hspace{0.2em}%
\begin{subfigure}{0.31\textwidth}
\includegraphics[width=\textwidth]{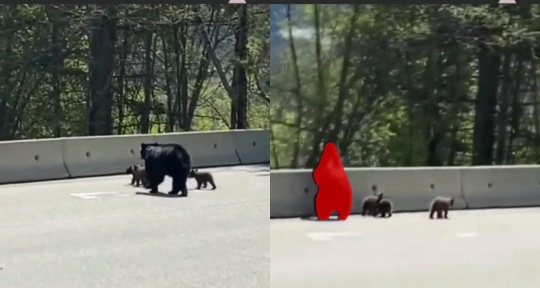}
\end{subfigure}}

%\resizebox{0.8\textwidth}{!}{
%\begin{subfigure}{0.31\textwidth}
%\includegraphics[width=\textwidth]{images/mocentric_qual_ablation_final/750/vidglamm/00020.jpg}
%\end{subfigure}%
%\hspace{0.2em}%
%\begin{subfigure}{0.31\textwidth}
%\includegraphics[width=\textwidth]{images/mocentric_qual_ablation_final/750/qwen/00020.jpg}
%\end{subfigure}%
%\hspace{0.2em}%
%\begin{subfigure}{0.31\textwidth}
%\includegraphics[width=\textwidth]{images/mocentric_qual_ablation_final/750/sa2va/00020.jpg}
%\end{subfigure}}

\vspace{1em}
\resizebox{0.8\textwidth}{!}{
\begin{subfigure}{0.31\textwidth}
\includegraphics[width=\textwidth]{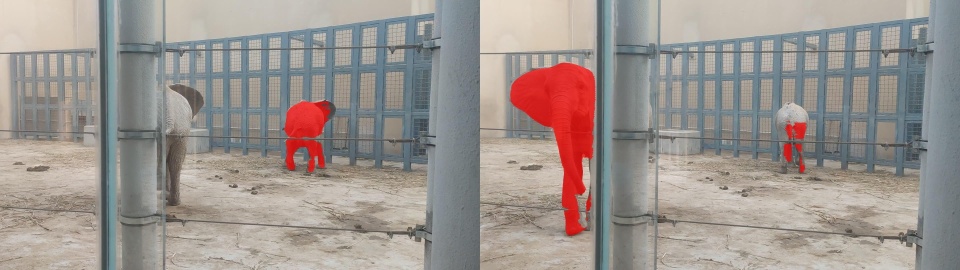}
\end{subfigure}%
\hspace{0.2em}%
\begin{subfigure}{0.31\textwidth}
\includegraphics[width=\textwidth]{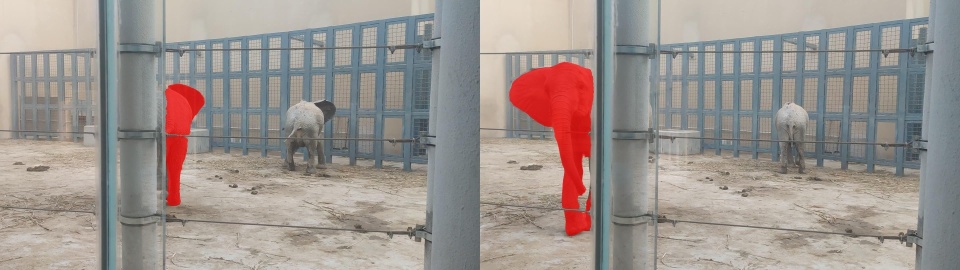}
\end{subfigure}%
\hspace{0.2em}%
\begin{subfigure}{0.31\textwidth}
\includegraphics[width=\textwidth]{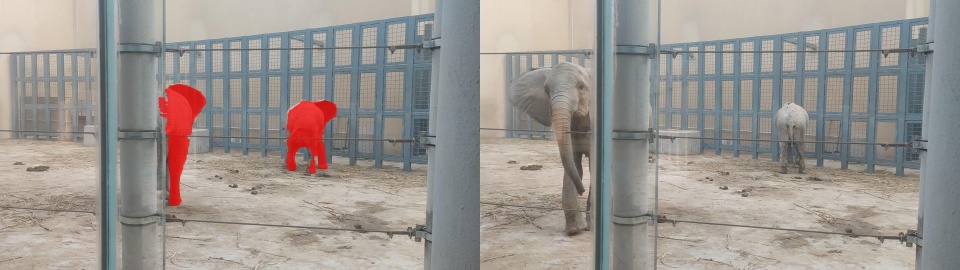}
\end{subfigure}}

\resizebox{0.8\textwidth}{!}{
\begin{subfigure}{0.31\textwidth}
\includegraphics[width=\textwidth]{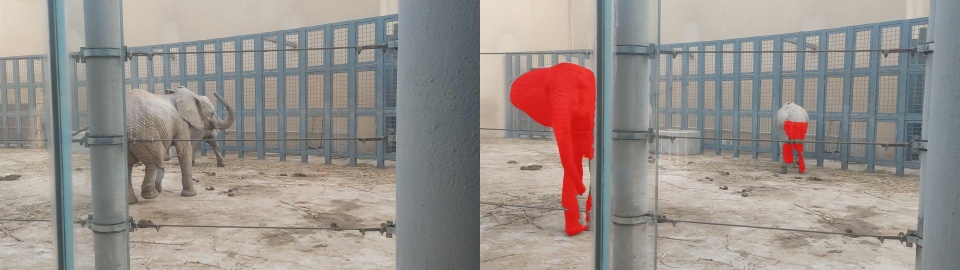}
\caption*{Video GLAMM}
\end{subfigure}%
\hspace{0.2em}%
\begin{subfigure}{0.31\textwidth}
\includegraphics[width=\textwidth]{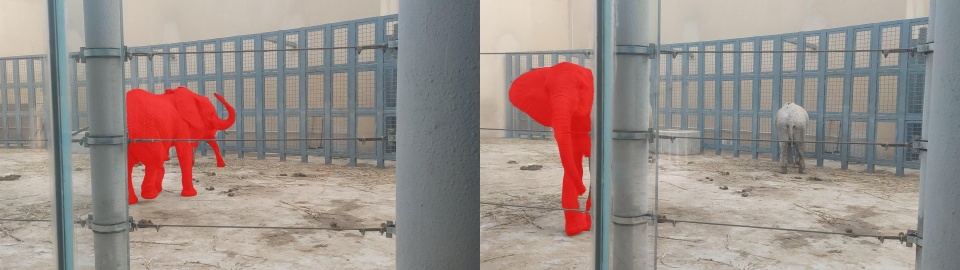}
\caption*{Baseline$\dagger$}
\end{subfigure}%
\hspace{0.2em}%
\begin{subfigure}{0.31\textwidth}
\includegraphics[width=\textwidth]{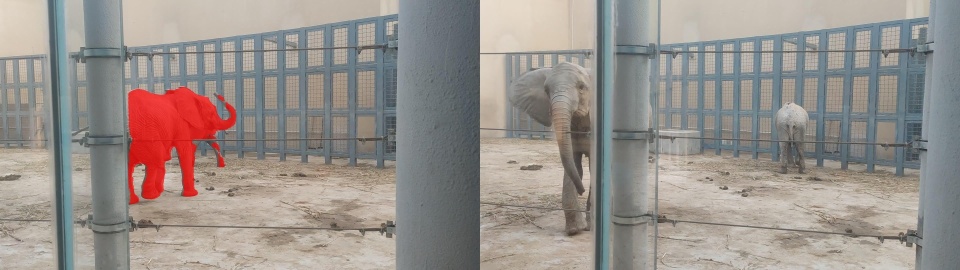}
\caption*{Sa2VA$\star$}
\end{subfigure}}

%\resizebox{0.8\textwidth}{!}{
%\begin{subfigure}{0.31\textwidth}
%\includegraphics[width=\textwidth]{images/mocentric_qual_ablation_final/d56_11/vidglamm/00040.jpg}
%\end{subfigure}%
%\hspace{0.2em}%
%\begin{subfigure}{0.31\textwidth}
%\includegraphics[width=\textwidth]{images/mocentric_qual_ablation_final/d56_11/qwen/00040.jpg}
%\end{subfigure}%
%\hspace{0.2em}%
%\begin{subfigure}{0.31\textwidth}
%\includegraphics[width=\textwidth]{images/mocentric_qual_ablation_final/d56_11/sa2va/00040.jpg}
%\end{subfigure}}
%\vspace{-0.5em}
%\vspace{-1em}
\caption{\textbf{Qualitative ablation on MoCentric-Bench} comparing our strongest baseline, Baseline$\dagger$, our adapted motion-centric model, Sa2VA$\star$ and concurrent work from VideoGLAMM. \textbf{Top rows:} \textit{val$^u$ \& Reverse}, expression is ``bear landing from a jump backward over a barricade and walking backward''. \textbf{Bottom rows:} \textit{val$^u$ \& Single frame}, expression is ``the elephant walking to the left''.}
\label{fig:qual_ablation}
\vspace{-1.5em}
\end{figure*}

\noindent \textbf{Qualitative analysis.} 
%Figure~\ref{fig:qual} shows the qualitative analysis for our strongest baseline, MLLM + SAM 2.0$\dagger$. The strongest baseline tends to segment the objects based on static cues in the referred expression. Hence, in the example provided ``elephants'' were segmented regardless of their motion, where the right-hand side has the static keyframe without any motion. Furthermore, 
We show a qualitative ablation of our strongest baseline and our motion-centric adapted models, Sa2VA$\star$, compared to previous work, VideoGLAMM, in Figure~\ref{fig:qual_ablation} on MoCentric-Bench (\textit{val$^u$ \& Single frame} and \textit{val$^u$ \& Reverse}). It shows two example sequences with two frames each. Our motion-centric adapted model, Sa2VA$\star$, has a better ability to differentiate the static frame from the dynamic video and, in certain cases, it can differentiate the forward from the reverse motion better than other methods. 

\begin{table}[t]
\centering
\caption{\textbf{Comparison of recent video MLLMs} to our single-image baselines, our motion-centric Sa2VA$\star$ and Molmo 2.0$\star$, on RefDAVIS'17 and MeVIS \textit{val} subset. Our baselines are mostly biased to the static information, yet surpass previously published works. $\dagger$: using automatic keyframe selection. $\star$: the motion-centric adaptation. Best and second-best are bolded and underlined, resp.}
\resizebox{0.95\textwidth}{!}{
\vspace{-0.5em}
\begin{tabular}{l|c|c|ccc|ccc}
\toprule
Method & Venue & Backbone/Base MLLM & \multicolumn{3}{c|}{RefDAVIS-17} & \multicolumn{3}{c}{MeVIS} \\
 &  & & $\mathcal{J}$ & $\mathcal{F}$ & $\mathcal{J\&F}$ & $\mathcal{J}$ & $\mathcal{F}$ & $\mathcal{J\&F}$ \\ \midrule
%  \multicolumn{9}{l}{\textbf{Closed Source MLLMs}} \\ \midrule
GPT-5 &  - &  - & - & - & 25.2 & - & - & 23.4 \\  \midrule
%Gemini-3 Pro &  - &  - & - & - & 66.6 & - & - & 42.5  \\ \midrule
 %\multicolumn{9}{l}{\textbf{Open Source Video/Image MLLMs}} \\ \midrule
ReferFormer & CVPR 2022 & VideoSwin-B & 58.1 & 64.1 & 61.1 & 29.8 & 32.2 & 31.0\\
LMPM & ICCV 2023 & Swin-T & - & - & - & 34.2 & 40.2 & 37.2 \\
LISA & CVPR 2024 & LLaVA-7B & 62.2 & 67.3 & 64.8 & 35.1 & 39.4 & 37.2 \\
TrackGPT& Arxiv 2023 & LLaVA-7B & 63.2 & 59.4 & 67.0  & 37.6 & 42.6  & 40.1 \\
VISA & ECCV 2024 & ChUniVi-7B & 66.3 & 72.5 & 69.4 & 40.7 & 46.3 & 43.5 \\
VidGLAMM & CVPR 2025 & VE+Phi3-mini-3.8B & 65.6 & 73.3 & 69.5 & 42.1 & 48.2 & 45.2 \\
RGA & ICCV 2025 & Qwen2.5-VL-7B & 68.3 & 77.3 & 72.8 & 47.4 & 52.8 & 50.1 \\
Sa2VA & TPAMI 2026 & InternVL2-8B & - & - & 75.2 & - & - & 46.9 \\
Sa2VA & TPAMI 2026 & InternVL2.5-8B & - & - & \underline{75.9} & - & - & 51.5 \\ 
Molmo 2.0 + SAM 2.0 & CVPR 2026 & Qwen3-8B & - & - & 72.7 & - & - & \textbf{62.3} \\ \midrule
Baseline & ours (NeurIPS 2026) & Qwen2.5-VL-7B & 67.4 & 73.5 & 70.5 & 41.3 & 47.7 & 44.5 \\ 
Baseline$\dagger$ & ours (NeurIPS 2026) & Qwen2.5-VL-7B & \underline{68.7} & \underline{74.7} & 71.7 & 43.5 & 50.2 & 46.9 \\ 
%Sa2VA $\star$ & (ours) 2025 & IVL2.5-8B & 71.7 & 80.1 & 75.9 & 48.3 & 55.6 & 51.9 \\
%Sa2VA $\star\star$ & (ours) 2025 & IVL2.5-8B & 72.0 & 81.3 & 76.7 & 46.8 & 54.0 & 50.4 \\
Sa2VA$\star$ & ours (NeurIPS 2026) & InternVL2.5-8B & \textbf{71.9} & \textbf{80.7} & \textbf{76.3} &  47.6 & 54.8 & 51.2 \\ 
Molmo 2.0$\star$ + SAM 2.0 & ours (NeurIPS 2026) & Qwen3-8B & 66.6 & 76.4 & 71.5 & \textbf{54.4} & \textbf{61.5} & \underline{58.0}\footnotemark \\ \bottomrule 
%%%Sa2VA$\star$ -  TS LR & (ours) 2025 & InternVL2.5-8B~\cite{chen2024expanding} & 69.9 & 78.8 & 74.3 &  &  & \\ 
%%%Sa2VA$\star$ -  TS LR All Data & (ours) 2025 & InternVL2.5-8B~\cite{chen2024expanding} & 67.6  & 77.0 & 72.3 &  &  & \\ \bottomrule % TS (L\&R) only
\end{tabular}}
\label{table:soa}
\vspace{-1em}
\end{table}

\footnotetext{Due to a large-scale PhP vulnerability that impacted the MeVIS evaluation server, these numbers were retrieved directly from MeVIS team evaluating the results.} 

\noindent \textbf{OOD scenarios.} We evaluate on two out-of-distribution (OOD) settings to confirm the benefit from our motion-centric adaptation, both of which follow probe $\mathcal{P}_1$. The first setting uses a new layout that was not seen during the adaptation. We define layout, $l_4$, following, $ \bar{S}_i = ( ([\bar{I}^t_i, I^t_i]_V)_{t=1}^T, \bar{E}_i, ([\bar{M}^t_i, M^t_i]_V)_{t=1}^T)$, where, $[., .]_V$ is the vertical spatial concatenation. Furthermore, we evaluate on a dataset that was concurrently released with our work~\cite{clark2026molmo2}, and follow the layout $l_2$. The results in Tab.~\ref{table:ood} consistently show that our motion-centric adaptation is better even in OOD scenarios, confirming it's a substantive gain in motion understanding. It also shows that not only MeVIS but also Molmo2Track, the animal subset, suffers from the same issue, where the expressions can easily be detected from a single frame in the video. We refer to Sa2VA performance on the original subset at 74.3\% vs. 47.7\% on our Molmo2Track motion-centric variant.

\begin{figure*}[t]
%\centering
\begin{minipage}{0.47\textwidth}
\captionof{table}{\textbf{Out-of-distribution scenarios (OOD),} using a new \textit{Layout}, $l_4$, with MeVIS \textit{val$^u$ \& Single frame} not seen during adaptation and a new \textit{Dataset} with layout $l_2$, Molmo2Track \textit{Animal \& Single frame}.}
\resizebox{\textwidth}{!}{
\begin{tabular}{ll|ccc}
\toprule
OOD Setup & Method & $\mathcal{J}$ & $\mathcal{F}$ & $\mathcal{J \& F}$ \\ \midrule
\multirow{2}{*}{\textit{Layout}} & Sa2VA & 24.6 & 33.6 & 29.1 \\
& Sa2VA$\star$ & \textbf{26.5} & \textbf{35.8} & \textbf{31.2} \\  \midrule
\multirow{2}{*}{\textit{Dataset}} & Sa2VA & 41.8 & 53.7 & 47.7 \\
& Sa2VA$\star$ & \textbf{42.5} & \textbf{54.2} & \textbf{48.4} \\  \bottomrule
\end{tabular}}
\label{table:ood}
\end{minipage}%
\hfill%
\begin{minipage}{0.5\textwidth}
\resizebox{\textwidth}{!}{
\begin{tikzpicture}
    \begin{axis}[
        ybar,
        height=5.5cm,
        width=12cm,
        xtick={1,2,3,4,5,6},
        xticklabels={\scriptsize{Multi-Step Actions}, \scriptsize{Multi-Object}, \scriptsize{Category Name}, \scriptsize{Color}, \scriptsize{Shape}, \scriptsize{Heading/Location}},
        xlabel= Expression Properties,
        ylabel= Percentage,
        enlarge x limits={abs=0.5},
        % ---------------------------------------------------------------------
        % changes to get what you want
        % ---------------------------------------------------------------------
        ymin=0.1,
        % remove the `xticks`
       xtick style={
                % (I this key has to be prefixed by `/pgfplots`, because
                %  normally here are just expected tikz keys)
                /pgfplots/major tick length=0pt
            },
       % every axis/.append style={font=\tiny},
        % remove the `xticks`
        legend style={
			area legend,
			at={(0.5,1.15)},
			anchor=north,
			legend columns=2},
    ]
    
        \addplot [blue, fill=blue!30]coordinates { (1, 52.9 ) (2, 45.6) (3, 42.8) (4, 45.6)  (5, 40.4) (6, 39.6)  }; % noticed two errors typos here: (4, 49.3) (5, 35)
	\addplot [red, fill=red!30]coordinates { (1, 47.1)  (2, 54.4) (3, 57.2) (4, 54.4) (5, 59.6) (6, 60.4) };
	 
        \legend{
            Motion Group, Static Group
        }
    \end{axis}
\end{tikzpicture}
}
\caption{\textbf{Motion expressions analysis.} Studying the properties of the two groups of expressions. Multi-Object: Multi-Object Interactions.}
\label{fig:motion_analysis}
\end{minipage}
\vspace{-1.5em}
\end{figure*}

\subsection{Conventional Benchmarks Results}
In this section, we compare against recent video MLLMs, which are struggling in MoCentric-Bench, yet some of them show strong results on conventional benchmarks. This indicates the benefit of our benchmark beyond conventional settings that mainly capture static cues in referring expressions. Table~\ref{table:soa} shows results across two benchmarks, including the motion referring expression segmentation dataset. It clearly shows that the simple baseline, Baseline, which does not incorporate any temporal information in the identification of the referred expression, surpasses VideoGLAMM, which is one of the recent methods, on RefDAVIS17. Our strongest baseline, Baseline$\dagger$, outperforms VideoGLAMM on both benchmarks and is competitive with concurrent works. 

This confirms that our baselines, which perform visual grounding on a single frame, establish strong results that can be safely used for our motion-centric probing and evaluation. While our baselines rely on a stronger base multimodal large language model than some of the previous methods, they are only meant to motivate the motion-centric evaluation and identify their shortcomings on our proposed benchmark and the corresponding analysis of the referring expressions. Additionally, our adapted variants, Sa2VA$\star$ and Molmo 2.0$\star$, either slightly outperform the original model or provide competitive performance, yet their real benefit was shown on MoCentric-Bench.

\subsection{Analysis on the referring motion expressions} 
In this section, we show that our probing can be used to analyze not only the models but the dataset itself in terms of the motion-referring expressions. This can be used to study the inherent biases in motion/static information when training on these data. We use our strongest baseline, Baseline$\dagger$, and compute the false positives in the static frame within our motion-centric evaluation (i.e., \textit{val$^u$ \& Single frame}) per video and referred expression. We then split the expressions into two major groups: the ones that resulted in less than 2\% false positive rate in the static frame, as the ``Motion'' group, and the ones that resulted in higher than that, as the ``Static'' group. We observe that almost half of the expressions at 380-460 out of 793 are in the ``Static" group. 

We use GPT-4o to identify certain properties in each expression by prompting it with the following: \textit{``Does the following expression have \texttt{<Property>}: \texttt{<EXP>}?''}. We replace \texttt{<Property>} with the following: (i) a multi-step action, (ii) a multi-object interaction, (iii) an identifiable category, (iv) color, (v) shape and (vi) heading, direction or location. We show the percentage of expressions within each property that received a positive response from GPT-4o, highlighting the differences between both the motion and static groups in Figure~\ref{fig:motion_analysis}. It shows that the motion group is mainly differentiated from the static one by multi-step actions, while the static group is mostly dominated by identifying the category, color, shape and heading/location. Multi-step actions relate to higher-level dynamics, yet these expressions will be insufficient to assess the models' ability to capture local motion, unlike MoCentric-Bench, which is tailored for this setting.

\section{Conclusions} %0.5 page

In conclusion, we have shown the shortcomings in visual grounding methods and datasets when considering local motion. We propose a novel benchmark that is motion-centric to probe the models' ability to understand motion existence and its ordering. Moreover, we propose strong baselines and a motion-centric adaptation that surpasses prior methods. 
%Finally, we perform a fine-grained analysis that provides insights into what properties differentiate the referred expressions that are dynamic focused vs. the static ones.

\section*{Acknowledgements}
This project was supported by free API credits provided by OpenAI through the Researcher Access Program. I have concurrently held NSERC grants RGPIN-2024-04100, DGECR-2024-00468 and ALLRP 595973-24 till June 2025, yet none were used in funding this project. %This project was implemented solely by me with a personal workstation.% as a challenge of what one AI researcher can do with a limited budget to contribute to multimodal large language models research.
%\section*{References}

\bibliographystyle{references}
\bibliography{references}

@article{liu2024visual,
  title={Visual instruction tuning},
  author={Liu, Haotian and Li, Chunyuan and Wu, Qingyang and Lee, Yong Jae},
  journal={Advances in Neural Information Processing Systems},
  volume={36},
  year={2023/}
}

@inproceedings{buch2022revisiting,
  title={Revisiting the" video" in video-language understanding},
  author={Buch, Shyamal and Eyzaguirre, Crist{\'o}bal and Gaidon, Adrien and Wu, Jiajun and Fei-Fei, Li and Niebles, Juan Carlos},
  booktitle={Proceedings of the IEEE/CVF conference on computer vision and pattern recognition},
  pages={2917--2927},
  year={2022}
}

@inproceedings{liu2024improved,
  title={Improved baselines with visual instruction tuning},
  author={Liu, Haotian and Li, Chunyuan and Li, Yuheng and Lee, Yong Jae},
  booktitle={Proceedings of the IEEE/CVF Conference on Computer Vision and Pattern Recognition},
  pages={26296--26306},
  year={2024}
}

@article{pont20172017,
  title={The 2017 davis challenge on video object segmentation},
  author={Pont-Tuset, Jordi and Perazzi, Federico and Caelles, Sergi and Arbel{\'a}ez, Pablo and Sorkine-Hornung, Alex and Van Gool, Luc},
  journal={arXiv preprint arXiv:1704.00675},
  year={2017}
}

@inproceedings{wang2024internvid,
  title={Internvid: A large-scale video-text dataset for multimodal understanding and generation},
  author={Wang, Yi and He, Yinan and Li, Yizhuo and Li, Kunchang and Yu, Jiashuo and Ma, Xin and Li, Xinhao and Chen, Guo and Chen, Xinyuan and Wang, Yaohui and others},
  booktitle={International Conference on Learning Representations},
  volume={2024},
  pages={42055--42079},
  year={2024}
}

@article{karim2023med++,
  title={MED-VT++: Unifying Multimodal Learning with a Multiscale Encoder-Decoder Video Transformer},
  author={Karim, Rezaul and Zhao, He and Wildes, Richard P and Siam, Mennatullah},
  journal={arXiv preprint arXiv:2304.05930},
  year={2023}
}

@inproceedings {molmo,
author = { Deitke, Matt and Clark, Christopher and Lee, Sangho and Tripathi, Rohun and Yang, Yue and Park, Jae Sung and Salehi, Mohammadreza and Muennighoff, Niklas and Lo, Kyle and Soldaini, Luca and Lu, Jiasen and Anderson, Taira and Bransom, Erin and Ehsani, Kiana and Ngo, Huong and Chen, YenSung and Patel, Ajay and Yatskar, Mark and Callison-Burch, Chris and Head, Andrew and Hendrix, Rose and Bastani, Favyen and VanderBilt, Eli and Lambert, Nathan and Chou, Yvonne and Chheda, Arnavi and Sparks, Jenna and Skjonsberg, Sam and Schmitz, Michael and Sarnat, Aaron and Bischoff, Byron and Walsh, Pete and Newell, Chris and Wolters, Piper and Gupta, Tanmay and Zeng, Kuo-Hao and Borchardt, Jon and Groeneveld, Dirk and Nam, Crystal and Lebrecht, Sophie and Wittlif, Caitlin and Schoenick, Carissa and Michel, Oscar and Krishna, Ranjay and Weihs, Luca and Smith, Noah A. and Hajishirzi, Hannaneh and Girshick, Ross and Farhadi, Ali and Kembhavi, Aniruddha },
booktitle = { 2025 IEEE/CVF Conference on Computer Vision and Pattern Recognition (CVPR) },
title = {{ Molmo and PixMo: Open Weights and Open Data for State-of-the-Art Vision-Language Models }},
year = {2025},
pages = {91-104}
}

@article{ding2025mevis,
  title={MeViS: A multi-modal dataset for referring motion expression video segmentation},
  author={Ding, Henghui and Liu, Chang and He, Shuting and Ying, Kaining and Jiang, Xudong and Loy, Chen Change and Jiang, Yu-Gang},
  journal={IEEE Transactions on Pattern Analysis and Machine Intelligence},
  year={2025},
  publisher={IEEE}
}

@article{bai2024one,
  title={One token to seg them all: Language instructed reasoning segmentation in videos},
  author={Bai, Zechen and He, Tong and Mei, Haiyang and Wang, Pichao and Gao, Ziteng and Chen, Joya and Liu, Lei and Zhang, Zheng and Shou, Mike Z},
  journal={Advances in Neural Information Processing Systems},
  volume={37},
  pages={6833--6859},
  year={2024}
}

@inproceedings{ilic2022appearance,
  title={Is appearance free action recognition possible?},
  author={Ilic, Filip and Pock, Thomas and Wildes, Richard P},
  booktitle={European Conference on Computer Vision},
  pages={156--173},
  year={2022},
  organization={Springer}
}

@inproceedings{clark2026molmo2,
  title={Molmo2: Open weights and data for vision-language models with video understanding and grounding},
  author={Clark, Christopher and Zhang, Jieyu and Ma, Zixian and Park, Jae Sung and Tripathi, Rohun and Lee, Sangho and Salehi, Mohammadreza and Ren, Jason and Kim, Chris Dongjoo and Yang, Yinuo and others},
  booktitle={Proceedings of the IEEE/CVF Conference on Computer Vision and Pattern Recognition},
  pages={28652--28668},
  year={2026}
}

@article{ghosh2023geneval,
  title={Geneval: An object-focused framework for evaluating text-to-image alignment},
  author={Ghosh, Dhruba and Hajishirzi, Hannaneh and Schmidt, Ludwig},
  journal={Advances in Neural Information Processing Systems},
  volume={36},
  pages={52132--52152},
  year={2023}
}

@inproceedings{kowal2022deeper,
  title={A deeper dive into what deep spatiotemporal networks encode: Quantifying static vs. dynamic information},
  author={Kowal, Matthew and Siam, Mennatullah and Islam, Md Amirul and Bruce, Neil DB and Wildes, Richard P and Derpanis, Konstantinos G},
  booktitle={Proceedings of the IEEE/CVF Conference on computer vision and pattern recognition},
  pages={13999--14009},
  year={2022}
}

@inproceedings{khoreva2018video,
  title={Video object segmentation with language referring expressions},
  author={Khoreva, Anna and Rohrbach, Anna and Schiele, Bernt},
  booktitle={Asian conference on computer vision},
  pages={123--141},
  year={2018},
  organization={Springer}
}

@inproceedings{li2018resound,
  title={Resound: Towards action recognition without representation bias},
  author={Li, Yingwei and Li, Yi and Vasconcelos, Nuno},
  booktitle={Proceedings of the European conference on computer vision (ECCV)},
  pages={513--528},
  year={2018}
}

@inproceedings{Perazzi2016,
  author = {F. Perazzi and J. Pont-Tuset and B. McWilliams and L. {Van Gool} and M. Gross and A. Sorkine-Hornung},
  title = {A Benchmark Dataset and Evaluation Methodology for Video Object Segmentation},
  booktitle = {Computer Vision and Pattern Recognition},
  year = {2016}
}

@article{
  siam2025pixfoundation,
  title={PixFoundation: Are We Heading in the Right Direction with Pixel-level Vision Foundation Models?},
  author={Mennatullah Siam},
  journal={Transactions on Machine Learning Research},
  issn={2835-8856},
  year={2026},
  url={https://openreview.net/forum?id=HNtjTTAac1},
}

@article{liu2023visual,
  title={Visual instruction tuning},
  author={Liu, Haotian and Li, Chunyuan and Wu, Qingyang and Lee, Yong Jae},
  journal={Advances in neural information processing systems},
  volume={36},
  pages={34892--34916},
  year={2023}
}

@article{zhu2023tracking,
  title={Tracking with human-intent reasoning},
  author={Zhu, Jiawen and Cheng, Zhi-Qi and He, Jun-Yan and Li, Chenyang and Luo, Bin and Lu, Huchuan and Geng, Yifeng and Xie, Xuansong},
  journal={arXiv preprint arXiv:2312.17448},
  year={2023}
}

@article{yuan2025sa2va,
  title={Sa2VA: Marrying SAM2 with MLLM for Dense Grounded Understanding of Images and Videos},
  author={Yuan, Haobo and Li, Xiangtai and Zhang, Tao and Sun, Yueyi and Huang, Zilong and Xu, Shilin and Ji, Shunping and Tong, Yunhai and Qi, Lu and Feng, Jiashi and Yang, Ming-Hsuan},
  journal={IEEE TPAMI},
  year={2026}
}

@article{kowal2024quantifying,
  title={Quantifying and learning static vs. dynamic information in deep spatiotemporal networks},
  author={Kowal, Matthew and Siam, Mennatullah and Islam, Md Amirul and Bruce, Neil DB and Wildes, Richard P and Derpanis, Konstantinos G},
  journal={IEEE Transactions on Pattern Analysis and Machine Intelligence},
  year={2024},
  publisher={IEEE}
}

@article{wang2025object,
  title={Object-centric Video Question Answering with Visual Grounding and Referring},
  author={Wang, Haochen and Chen, Qirui and Yan, Cilin and Cai, Jiayin and Jiang, Xiaolong and Hu, Yao and Xie, Weidi and Gavves, Stratis},
  journal={arXiv preprint arXiv:2507.19599},
  year={2025}
}

@inproceedings{lai2024lisa,
  title={Lisa: Reasoning segmentation via large language model},
  author={Lai, Xin and Tian, Zhuotao and Chen, Yukang and Li, Yanwei and Yuan, Yuhui and Liu, Shu and Jia, Jiaya},
  booktitle={Proceedings of the IEEE/CVF Conference on Computer Vision and Pattern Recognition},
  pages={9579--9589},
  year={2024}
}

@inproceedings{kazemzadeh2014referitgame,
  title={Referitgame: Referring to objects in photographs of natural scenes},
  author={Kazemzadeh, Sahar and Ordonez, Vicente and Matten, Mark and Berg, Tamara},
  booktitle={Proceedings of the Conference on Empirical Methods in Natural Language Processing (EMNLP)},
  pages={787--798},
  year={2014}
}

@inproceedings{zhang2025llava,
  title={Llava-grounding: Grounded visual chat with large multimodal models},
  author={Zhang, Hao and Li, Hongyang and Li, Feng and Ren, Tianhe and Zou, Xueyan and Liu, Shilong and Huang, Shijia and Gao, Jianfeng and Li, Chunyuan and Yang, Jainwei and others},
  booktitle={Proceedings of the European Conference on Computer Vision},
  pages={19--35},
  year={2024},
  organization={Springer}
}

@inproceedings{
ravi2024sam,
title={{SAM} 2: Segment Anything in Images and Videos},
author={Nikhila Ravi and Valentin Gabeur and Yuan-Ting Hu and Ronghang Hu and Chaitanya Ryali and Tengyu Ma and Haitham Khedr and Roman R{\"a}dle and Chloe Rolland and Laura Gustafson and Eric Mintun and Junting Pan and Kalyan Vasudev Alwala and Nicolas Carion and Chao-Yuan Wu and Ross Girshick and Piotr Dollar and Christoph Feichtenhofer},
booktitle={The Thirteenth International Conference on Learning Representations},
year={2025},
}

@inproceedings{munasinghe2024videoglamm,
  title        = {VideoGLaMM: A Large Multimodal Model for Pixel‐Level Visual Grounding in Videos},
  author       = {Munasinghe, Shehan and Gani, Hanan and Zhu, Wenqi and Cao, Jiale and Xing, Eric and Khan, Fahad Shahbaz and Khan, Salman},
  booktitle    = {Proceedings of the IEEE/CVF Conference on Computer Vision and Pattern Recognition (CVPR)},
  year         = {2025},
  pages        = {19036--19046}, 
}

@inproceedings{llin2023videoo,
    title = {Video-{LL}a{VA}: Learning United Visual Representation by Alignment Before Projection},
    author = {Lin, Bin  and Ye, Yang  and Zhu, Bin  and Cui, Jiaxi  and Ning, Munan  and Jin, Peng  and Yuan, Li},
    booktitle = {Proceedings of the 2024 Conference on Empirical Methods in Natural Language Processing},
    year = {2024}
}

@article{fang2024mmbench,
  title={Mmbench-video: A long-form multi-shot benchmark for holistic video understanding},
  author={Fang, Xinyu and Mao, Kangrui and Duan, Haodong and Zhao, Xiangyu and Li, Yining and Lin, Dahua and Chen, Kai},
  journal={Advances in Neural Information Processing Systems},
  volume={37},
  pages={89098--89124},
  year={2024}
}

@inproceedings{fu2024video,
  author       = {Chaoyou Fu and
                  Yuhan Dai and
                  Yongdong Luo and
                  Lei Li and
                  Shuhuai Ren and
                  Renrui Zhang and
                  Zihan Wang and
                  Chenyu Zhou and
                  Yunhang Shen and
                  Mengdan Zhang and
                  Peixian Chen and
                  Yanwei Li and
                  Shaohui Lin and
                  Sirui Zhao and
                  Ke Li and
                  Tong Xu and
                  Xiawu Zheng and
                  Enhong Chen and
                  Caifeng Shan and
                  Ran He and
                  Xing Sun},
  title        = {Video-MME: The First-Ever Comprehensive Evaluation Benchmark of Multi-modal
                  LLMs in Video Analysis},
  booktitle    = {{IEEE/CVF} Conference on Computer Vision and Pattern Recognition,
                  {CVPR}},
  pages        = {24108--24118},
  year         = {2025},
}

@inproceedings{miao2021vspw,
  title={Vspw: A large-scale dataset for video scene parsing in the wild},
  author={Miao, Jiaxu and Wei, Yunchao and Wu, Yu and Liang, Chen and Li, Guangrui and Yang, Yi},
  booktitle={Proceedings of the IEEE/CVF conference on computer vision and pattern recognition},
  pages={4133--4143},
  year={2021}
}

@inproceedings{zohar2024apollo,
  author={Zohar, Orr and Wang, Xiaohan and Dubois, Yann and Mehta, Nikhil and Xiao, Tong and Hansen-Estruch, Philippe and Yu, Licheng and Wang, Xiaofang and Juefei-Xu, Felix and Zhang, Ning and Yeung-Levy, Serena and Xia, Xide},
  booktitle={IEEE/CVF Conference on Computer Vision and Pattern Recognition (CVPR)}, 
  title={Apollo: An Exploration of Video Understanding in Large Multimodal Models}, 
  year={2025},
  pages={18891-18901}}

@article{choi2019can,
  title={Why can't i dance in the mall? learning to mitigate scene bias in action recognition},
  author={Choi, Jinwoo and Gao, Chen and Messou, Joseph CE and Huang, Jia-Bin},
  journal={Advances in Neural Information Processing Systems},
  volume={32},
  year={2019}
}

@misc{cores2024tvbench,
  author = {Daniel Cores and Michael Dorkenwald and Manuel Mucientes and Cees G. M. Snoek and Yuki M. Asano},
  title = {TVBench: Redesigning Video-Language Evaluation},
  year = {2024},
  eprint = {arXiv:2410.07752},
}

@article{xue2025seeing,
  title={Seeing the arrow of time in large multimodal models},
  author={Xue, Zihui Sherry and Luo, Romy and Grauman, Kristen},
  journal={Advances in Neural Information Processing Systems},
  volume={38},
  pages={90925--90955},
  year={2025}
}

@inproceedings{maaz2023video,
    title = "Video-{C}hat{GPT}: Towards Detailed Video Understanding via Large Vision and Language Models",
    author = "Maaz, Muhammad  and
      Rasheed, Hanoona  and
      Khan, Salman  and
      Khan, Fahad",
    booktitle = "Proceedings of the 62nd Annual Meeting of the Association for Computational Linguistics (Volume 1: Long Papers)",
    year = "2024",
}

@inproceedings{ding2023mevis,
  title={MeViS: A large-scale benchmark for video segmentation with motion expressions},
  author={Ding, Henghui and Liu, Chang and He, Shuting and Jiang, Xudong and Loy, Chen Change},
  booktitle={Proceedings of the IEEE/CVF international conference on computer vision},
  pages={2694--2703},
  year={2023}
}

@article{bai2023qwen,
  title={Qwen technical report},
  author={Bai, Jinze and Bai, Shuai and Chu, Yunfei and Cui, Zeyu and Dang, Kai and Deng, Xiaodong and Fan, Yang and Ge, Wenbin and Han, Yu and Huang, Fei and others},
  journal={arXiv preprint arXiv:2309.16609},
  year={2023}
}

@article{bai2025qwen2,
  title={Qwen2. 5-vl technical report},
  author={Bai, Shuai and Chen, Keqin and Liu, Xuejing and Wang, Jialin and Ge, Wenbin and Song, Sibo and Dang, Kai and Wang, Peng and Wang, Shijie and Tang, Jun and others},
  journal={arXiv preprint arXiv:2502.13923},
  year={2025}
}

@inproceedings{szot2025multimodal,
  title={From multimodal llms to generalist embodied agents: Methods and lessons},
  author={Szot, Andrew and Mazoure, Bogdan and Attia, Omar and Timofeev, Aleksei and Agrawal, Harsh and Hjelm, Devon and Gan, Zhe and Kira, Zsolt and Toshev, Alexander},
  booktitle={Proceedings of the Computer Vision and Pattern Recognition Conference},
  pages={10644--10655},
  year={2025}
}

@article{zhu2025internvl3,
  title={InternVL3: Exploring Advanced Training and Test-Time Recipes for Open-Source Multimodal Models},
  author={Zhu, Jinguo and Wang, Weiyun and Chen, Zhe and Liu, Zhaoyang and Ye, Shenglong and Gu, Lixin and Duan, Yuchen and Tian, Hao and Su, Weijie and Shao, Jie and others},
  journal={arXiv preprint arXiv:2504.10479},
  year={2025}
}

@article{chen2024expanding,
  title={Expanding performance boundaries of open-source multimodal models with model, data, and test-time scaling},
  author={Chen, Zhe and Wang, Weiyun and Cao, Yue and Liu, Yangzhou and Gao, Zhangwei and Cui, Erfei and Zhu, Jinguo and Ye, Shenglong and Tian, Hao and Liu, Zhaoyang and others},
  journal={arXiv preprint arXiv:2412.05271},
  year={2024}
}

@inproceedings{yan2024visa,
  title={Visa: Reasoning video object segmentation via large language models},
  author={Yan, Cilin and Wang, Haochen and Yan, Shilin and Jiang, Xiaolong and Hu, Yao and Kang, Guoliang and Xie, Weidi and Gavves, Efstratios},
  booktitle={European Conference on Computer Vision},
  pages={98--115},
  year={2024},
  organization={Springer}
}

@inproceedings{seo2020urvos,
  title={Urvos: Unified referring video object segmentation network with a large-scale benchmark},
  author={Seo, Seonguk and Lee, Joon-Young and Han, Bohyung},
  booktitle={European conference on computer vision},
  pages={208--223},
  year={2020},
  organization={Springer}
}

@inproceedings{chen2024internvl,
  title={Internvl: Scaling up vision foundation models and aligning for generic visual-linguistic tasks},
  author={Chen, Zhe and Wu, Jiannan and Wang, Wenhai and Su, Weijie and Chen, Guo and Xing, Sen and Zhong, Muyan and Zhang, Qinglong and Zhu, Xizhou and Lu, Lewei and others},
  booktitle={Proceedings of the IEEE/CVF conference on computer vision and pattern recognition},
  pages={24185--24198},
  year={2024}
}

@inproceedings{wu2022language,
  title={Language as queries for referring video object segmentation},
  author={Wu, Jiannan and Jiang, Yi and Sun, Peize and Yuan, Zehuan and Luo, Ping},
  booktitle={Proceedings of the IEEE/CVF Conference on Computer Vision and Pattern Recognition},
  pages={4974--4984},
  year={2022}
}

@article{karim2023med,
  title={MED-VT++: Unifying Multimodal Learning with a Multiscale Encoder-Decoder Video Transformer},
  author={Karim, Rezaul and Zhao, He and Wildes, Richard P and Siam, Mennatullah},
  journal={arXiv preprint arXiv:2304.05930},
  year={2023}
}

@inproceedings{
hu2022lora,
title={Lo{RA}: Low-Rank Adaptation of Large Language Models},
author={Edward J Hu and Yelong Shen and Phillip Wallis and Zeyuan Allen-Zhu and Yuanzhi Li and Shean Wang and Lu Wang and Weizhu Chen},
booktitle={International Conference on Learning Representations},
year={2022}
}

%%%%%%%%%%%%%%%%%%%%%%%%%%%%%%%%%%%%%%%%%%%%%%%%%%%%%%%%%%%%
\newpage

\appendix

%\iffalse
%\section{Supplemental Video}
%\label{sec:supp_video}
%We include an accompanying 4-minute video as part of the supplemental materials. In this video, we highlight our three contributions, where our motion-centric probing and evaluation quantify failures of the benchmarks in emphasizing motion within the motion referring expressions, in addition to quantifying the failures of recent video-MLLMs (e.g, Sa2VA) in identifying local motion patterns. This is followed by successful examples of our motion-centric adaptation and failure cases.
%\fi

\begin{table}[t]
\centering
\caption{Detailed statistics of our synthetically generated dataset.}
\resizebox{0.95\textwidth}{!}{
\begin{tabular}{l|c|c|c|c}
\toprule
\textbf{Dataset Source} &  \textbf{Variant}      &  \textbf{Videos} & \textbf{Probe \& Layout} & \textbf{Video-Expression pairs} \\ \midrule
MeVIS (train)               & MoCentric Train        & 3,324             & $\mathcal{P}_1$, ($l_2, l_3$)             & 46,102                          \\ 
MeVIS (val$^u$)               & val$^u$ \& Single frame   & 150               & $\mathcal{P}_1$, ($l_2, l_3, l_4$)        & 2,379                          \\ 
MeVIS (val$^u$)                 & val$^u$ \& Blank image    & 100               &  $\mathcal{P}_3$, ($l_2, l_3$)             & 1,586                           \\
 MeVIS (val$^u$)                 &  Reverse                & 28                &  $\mathcal{P}_2$, ($l_1$)                 & 152                            \\ 
 MeVIS (val$^u$)                   &  val$^u$ \& Reverse        & 56                & $\mathcal{P}_2$, ($l_2, l_3$)            &  304                            \\ 
 Molmo2Track (Animal)    & Animal \& Single frame & 188              & $\mathcal{P}_1$, ($l_2$)               & 332                             \\ \midrule
Total                  &  -                      & 3,846       & -                        & $\sim$50,800                   \\ \bottomrule
\end{tabular}}
\vspace{1em}
\label{tab:mocentric_bench_stats_detailed}
\end{table}

\begin{table*}[t]
\caption{\textbf{Our synthetic benchmark \textit{val$^u$ \& Reverse}.} Examples of the original motion expression for the forward motion and its reverse motion expression for the reverse video.}
\centering
%\resizebox{0.49\textwidth}{!}{
%\begin{tabular}{l|ll}
\begin{tabular}{l|p{60mm}|p{60mm}}
\toprule
Exp. \# & Forward Motion Expression & Reverse Motion Expression \\ \midrule
1& turn and walk away from us & turn and walk towards us \\
2& the elephant walking to the left & the elephant walking to the right \\
3& the moving pair of elephants. & the pair of reversing elephants \\
4& the individual providing food to the lizard & the individual taking food away from the lizard \\
5& the elephant pressing its nose against the other elephant's back & the elephant moving its nose away from the other elephant's back \\
6& the aircraft moving towards us & the aircrat moving away from us \\
7& Man pushing bicycle around in a circle & man pulling bicycle around in a circle in reverse \\
8 & the two model airplanes in motion & the two model aircraft in reverse movement \\
9 & dog walking then turn around & dog turning around then walking backward\\
10 & the male child going to embrace the small dog & the male child moving away from the small dog after an embrace \\ \bottomrule
\bottomrule
\end{tabular}%}
\label{table:reverse_expressions}
\end{table*}

% IoU_Bg the modified has a pitfall that its not capturing the false positives that are outside the designated fooreground region that is treated as background which reduce the differences :(! Sticking to the False positive rate for now
\section{Additional Implementation Details}
\label{sec:imp_details}

\textbf{Datasets.}
For the synthetic dataset, we use the HuggingFace ``Qwen2.5-VL-7B-Instruct'' weights for the benchmark construction to select the static keyframe, and GPT-4o for the automatic reversal. In MoCentric-Bench evaluation, the motion ordering probe, filtered out 641 video-expression pairs and relied on 152 pairs, since the former had the reverse expression grounding ambiguous with respect to the forward, in the video. In the motion existence probe, all the 793 video-expression pairs were used. Table~\ref{tab:mocentric_bench_stats_detailed} shows the breakdown of the dataset variants, sources and statistics. We add more examples from the \textit{val$^u$ \& Reverse} to show the reverse motion expressions automatically generated using GPT-4o in Table~\ref{table:reverse_expressions}. Moreover, in the appendix results, we create an additional probe to evaluate the spatial bias, $\mathcal{P}_3$, which creates a video of a black/white image with the respective empty masks and then combines it with the original video following $l_2, l_3$ layouts. We refer to this variant as \textit{val$^u$ \&  Blank image} and evaluate all our models on it to showcase the impact of the spatial bias, if any, without introducing any deceiving motion. Similar to the previous probes, we evaluate on $l_2$, yet we use both $l_2, l_3$ during the inference of our single-image baselines, then combine their predictions to evaluate the models' spatial biases for both layouts.

\noindent \textbf{Training.} For the motion-centric adaptation of Sa2VA, we follow a similar procedure to Sa2VA training, but instead we add the motion-centric synthetically generated MeVIS training variants following our motion probes and layouts. We use a dataset ratio of 1:1 during sampling from both the original MeVIS training set and our motion-centric synthetic one. Similarly, for the motion-centric adaptation of Molmo 2.0, we use MeVIS, Motion-Centric MeVIS, ReVOS and RefYTVOS datasets for both grounding and tracking as our training mixture, with a ratio of 1:1 between grounding and tracking tasks for 3,000 iterations. For Sa2VA, we set the LoRA finetuning hyperparameters with rank $r=128$ and $\alpha=256$, while for Molmo 2.0 we set these to $r=32$ and $\alpha=64$; this is conducted for the vision encoder. We use the trained Sa2VA weights as ``ByteDance/Sa2VA-8B'' that are provided in HuggingFace as the initialization for our model. Similarly, for Molmo 2.0, we use their weights ``allenai/Molmo2-8B'' as the initialization. In both, we finetune the vision encoder and projector with a learning rate of $4\times10^-5$. Throughout the experiments, we use a single A6000 GPU for the training and the evaluation of the models discussed.

\noindent \textbf{Inference.} 
For our single-image baselines, we use the following weights for Qwen2.5-VL that are available from HuggingFace ``Qwen2.5-VL-7B-Instruct''. For the motion-centric probing and to avoid any spatial bias that might exist in Qwen2.5-VL, we rather prompt the models with layout $l_2$ and $l_3$ for the same video, then we use the combination of both predictions for the SAM 2.0 initialization. Specifically, the bounding boxes from $l_3$ are reordered spatially to align with $l_2$ layout and the bounding boxes from both are used to initialize SAM 2.0. 

We use the following prompt to retrieve the bounding boxes from Qwen2.5-VL: \textit{``Locate the \texttt{<EXP>}, output its bounding box coordinates using JSON format.''}, with the first frame in the video. In case the model's JSON response can not be parsed into a bounding box, we allow the model to use the first frame from the beginning of the video that provides a valid output. Similarly, in MLLM + SAM 2.0 $\dagger$, if the output can not be parsed into a bounding box, it can use the first frame from the beginning of the temporal window that is automatically selected corresponding to the referred object, till a valid output is identified. That baseline can not capture the full spatiotemporal information in the video, since the pixel-level grounding only uses one static keyframe. 

For Sa2VA, we also report in the appendix the average predictions combining both $l_2, l_3$ layouts. Note that, unlike our single-image baselines, Sa2VA is trained end-to-end, combining both the MLLM and SAM 2.0; as such, we combine the average predictions from the final output of Sa2VA rather than combining before SAM 2.0 initialization as in our baselines. We reorder spatially the predictions from $l_3$ to align with $l_2$ layout, then average both. 

For LMPM and VideoGLAMM, we use their original implementation and hyperparameters as is. However, on the \textit{Single frame} evaluation, we observe that LMPM tends to favour the background class due to their high predefined threshold on the output predictions at 0.8; as such, for fairness with respect to other methods, we set it to 0.5 as standard only in the \textit{Single frame} evaluation. %notice the original implementation relies on a threshold for 

\noindent \textbf{Evaluation.} We use $\mathcal{J}$ and $\mathcal{F}$ metrics following the standard evaluation from video segmentation~\cite{Perazzi2016,pont20172017}. In the appendix, we also rely on the false positive rate, $\mathcal{FPR} = \frac{FP}{TN+FP}$, to evaluate the false positives in the deceiving region. The deceiving region in probe $\mathcal{P}_1$ corresponds to the repeated static keyframe (i.e., static side), while in probe $\mathcal{P}_2$ it corresponds to the original forward motion (i.e., forward side), since the reverse motion expressions are the ones used. The results for Sa2VA$\star$ are the average of two runs.

Additionally, we use a modified metric for the false positive rate, $\mathcal{FPR}_{\text{ffg}}$, that evaluates the false positives within the fake foreground in the deceiving region. Since it is not possible to accurately predict what would be considered as the fake foreground, that is, the areas that tend to mislead video MLLMs with respect to the referred expression, we rather use the ground-truth mask corresponding to the static keyframe or the forward motion according to the used motion probe. The goal for evaluating the false positive rate within the fake foreground, $\mathcal{FPR}_{\text{ffg}}$, is to demonstrate the failures without being dominated by background pixels in the full deceiving region conveyed as in $\mathcal{FPR}$. Nonetheless, $\mathcal{FPR}$ is more accurate as it evaluates the full region, unlike $\mathcal{FPR}_{\text{ffg}}$, which can miss false positives that are not within the identified fake foreground.

\section{Additional Ablation Studies}
\label{sec:more_results}

\textbf{Ablation of our motion-centric adaptation.} We provide an additional ablation that studies the failures in the deceiving region in our two probes, $\mathcal{P}_1, \mathcal{P}_2$, which correspond to the static keyframe in the former and the forward motion in the latter and ablating our motion-centric adaptation. Table~\ref{table:fpr_mocentric} confirms that our adaptation reduces the false positives in the deceiving region. Note that the false positive rate is dominated by the predicted background pixels as the true negatives; as such, the contour accuracy and region similarity metrics, $\mathcal{J} \& \mathcal{F}$, reflect better the performance difference. Furthermore, we compare the motion-centric variant, Molmo 2.0$\star$, with a baseline of finetuning the vision encoder and projector for Molmo 2.0 without using the motion-centric MeVIS variants in the training data mixture, evaluating $\mathcal{J}\&\mathcal{F}$ metrics on the \textit{val$^u$ \& Reverse} split. In both experiments, we use the same hyperparameters to ensure a fair comparison. Molmo 2.0$\star$ is at 38.1\% vs. the Molmo 2.0 baseline, which is at 32.9\%. Hence, additional finetuning of Molmo 2.0 without the use of our synthetic motion-centric MeVIS does not improve the model's ability to ground motion.

\noindent \textbf{Results for \textit{Single frame}.} We evaluate the methods on the \textit{Single frame} setup, although it has its pitfalls as it favours models that prioritize background, and it can easily be solved by a model that always predicts background. Nonetheless, it is a good evaluation to showcase the failures that current visual grounding methods exhibit with fake motion without the impact of any spatial biases. Table~\ref{table:single} shows the false positive rate, $\mathcal{FPR}$, on the deceiving region and its modified one that emphasizes the fake foreground areas, $\mathcal{FPR}_{\text{ffg}}$. It clearly shows there are failures across all the models that are majorly deceived with the regions that are static. Note that the $\mathcal{FPR}_{\text{ffg}}$ metric can miss false positives outside what we predefined as fake foreground, as such, the false positive rate $\mathcal{FPR}$ results in a more accurate evaluation that takes all the false positives into account. Surprisingly, we found in this specific setup that our single-image baselines exhibit lower false positive rates than all the other models, including Sa2VA and Sa2VA$\star$. Nonetheless, analyzing this result further, we notice this is due to the nature of the setup that favours models predicting all pixels as background.  We notice that our baselines, which take a single image as input when asked to detect the referred expression, might fail to generate valid bounding boxes across all the video and rather resort to predicting all the pixels as background. We observe that across the 793 pairs, around 40\% either show the aforementioned failure, which deceivingly gives a low false positive rate, yet it exhibits a failure, or it shows Sa2VA performing better in the false positive rate than our strongest baseline. 

\noindent \textbf{Spatial bias ablation.} We conduct two additional ablations to confirm that the failures we observe in our MoCentric-Bench stem from failures to capture local motion patterns aside from any spatial biases. First, we evaluate on the spatial bias probe, $\mathcal{P}_3$, that uses a blank single colour image to replace the static keyframe we used in our probe, $\mathcal{P}_1$. Table~\ref{table:black_flag} shows that the spatial bias probe introduces minimal degradation relative to our motion probe. For example, in VideoGLAMM, the spatial bias can result in less than 1\% degraded performance, while our motion probe results in more than 25\% degradation. Hence, around half the original performance is reduced by limitations in video MLLMs in capturing the local motion patterns, which is a safety concern.

Second, we report for both the original Sa2VA and our motion-centric adaptation their performance when combining predictions from both layouts $l_2, l_3$. Table~\ref{table:flip} confirms that the results are still within half the original performance of the respective models. Regardless of whether we use one layout, $l_2$, during the inference or combine both layouts' predictions to account for any spatial biases if they exist. It also confirms the benefit from our adaptation regardless of the evaluation protocol. Both ablations emphasize that the spatial bias has a minimal role in our conclusion that state-of-the-art video MLLMs show failures in understanding and identifying local motion and its interaction with language.

\begin{figure*}[t]
\begin{minipage}{0.48\textwidth}
\captionof{table}{\textbf{Ablation of the motion centric adaptation} on \textit{val$^u$ \& Single frame} and \textit{val$^u$ \& Reverse} MeVIS variants showing their average score evaluating on the whole video, $\mathcal{J} \& \mathcal{F}$, and the $\mathcal{FPR}$ in the static frame or the forward side only.}
\centering
\resizebox{0.73\textwidth}{!}{
\begin{tabular}{ll|cc}
\toprule
 Method &  $\mathcal{J} \& \mathcal{F} \uparrow$  &   $\mathcal{FPR}  \downarrow$ \\ \midrule
\multicolumn{3}{l}{\textit{val$^u$ \& Single frame}} \\ \midrule
Sa2VA & 28.5 & 7.3 \\
Sa2VA$\star$ &  \textbf{31.1} & \textbf{6.8} \\ \midrule
\multicolumn{3}{l}{\textit{val$^u$ \& Reverse}} \\ \midrule
Sa2VA & 37.0 & 6.5  \\
Sa2VA$\star$ & \textbf{42.0}  & \textbf{5.3} \\ \bottomrule
\end{tabular}}
\label{table:fpr_mocentric}
\end{minipage}%
\hfill%
\begin{minipage}{0.48\textwidth}
\captionof{table}{\textbf{Evaluation on the \textit{Single frame}} MeVIS variant showing the false positive rate, $\mathcal{FPR}$, and the modified one that is focused on the fake foreground, $\mathcal{FPR}_{\text{ffg}}$. This setup favours models that tend to predict background class.}
\centering
\resizebox{\textwidth}{!}{
\begin{tabular}{l|cc}
\toprule
Method & $\mathcal{FPR} \downarrow$ & $\mathcal{FPR}_{\text{ffg}} \downarrow$  \\ \midrule
%LMPM - $\mu=0.8$   & 31.2  \\
LMPM - thresh=0.5 & 12.9 & 73.2  \\ %- $\mu=0.5$ 
VideoGLAMM & 9.5 & 67.1 \\
RGA & 8.0 & 68.9 \\
Sa2VA  & 8.4 & 67.6  \\ \midrule
MLLM + SAM 2.0 & 7.1 & \textbf{57.6}  \\ 
MLLM + SAM 2.0$\dagger$ & \textbf{7.0} & 58.0 \\ 
%Sa2VA$\star$ - Run 1& 8.3 & 69.7 \\
%Sa2VA$\star$ - Run 2& 8.2 & 68.2 \\ 
Sa2VA$\star$  & 8.2 &69.0 \\ \bottomrule
%Sa2VA$\star\star$ - TS LR All Data & 8.3 & 68.9  \\ 
%Sa2VA$\star\star$ - TS+S All Data & &   \\ 
%Sa2VA$\star\star$ - TS LR & 7.7 & 62.7 \\ 
%Sa2VA$\star\star$ -  TS+S & & \\ \bottomrule
\end{tabular}}
%\label{table:single}
%\caption{Evaluation on the \textit{Single frame} MeVIS variant showing the mean intersection over union for the focused background class, $\mathcal{J}_{Bg}.}
\vspace{0.5em}
\label{table:single}
\end{minipage}
\end{figure*}

\textbf{Keyframe selection ablation.} Additionally, we provide an ablation study on the choice of the frame in our baselines, where we compare the selection of the first frame, which is our first baseline MLLM + SAM 2.0 vs. the last frame vs. the automatically selected keyframe, which is our second baseline MLLM + SAM 2.0 $\dagger$. While the keyframe selection is a design choice that does not impact our core contribution, we aim to provide the strongest baseline that relies on visual grounding from a single frame, which is improved by a better initialization of SAM 2.0. Table~\ref{table:app_ablations} shows that the keyframe selection is persistently better across all metrics with a considerable margin. Hence, we keep the keyframe selection as our strongest baseline.

\textbf{Removing static expressions.} We perform an additional ablation to confirm the degradation is related to capturing motion patterns in the motion existence probe, by removing any expression that describes a static position for the object. We use GPT-4o to inspect all MeVIS val$^u$ expressions whether they describe static positions or motion. We find that 148 out of 793 expressions pairs represent static expressions, e.g., 
``sitting still on the left without changing position, with side facing us'', ``the monkey squatting in the middle of the cave entrance'', ``the left one of the two sitting people in the distance'', ``the parked vehicle'', ``stationary bicycle''. We evaluate Sa2VA performance on \textit{val$^u$} and the motion-centric variant \textit{val$^u$ \& Single frame} in Table~\ref{table:nostatic}. It clearly shows that even when focusing on the remaining expressions that are only focused on describing moving objects, state-of-the-art video MLLMs still fail in differentiating the true vs. fake motion, resulting in degrading the peformance by half.

\begin{table}[t]
\centering
\caption{\textbf{Studying spatial biases.} Ablation on the impact of our motion probe, $\mathcal{P}_1$, in \textit{val$^u$ \& Single frame} variant vs. the spatial bias probe, $\mathcal{P}_3$, in \textit{val$^u$ \& Blank image} variant. $\Delta$: the difference between the original performance on MeVIS \textit{val$^u$} and the respective probe performance. It clearly shows that spatial bias has minimal impact relative to the impact of our motion probe, where models are degraded by almost half in the case of \textit{val$^u$ \& Single frame}. This is due to the models being misled by the fake motion from the static frame. As such, it confirms the limitations in the current benchmarks and video MLLMs for visual grounding in capturing local motion.}
\resizebox{0.75\textwidth}{!}{
\begin{tabular}{l|c|cc|cc}
\toprule
Method & \textit{val$^u$} & \multicolumn{2}{c|}{ \textit{val$^u$ \& Single frame}} & \multicolumn{2}{c}{ \textit{val$^u$ \& Blank image}}\\ 
& $\mathcal{F} \& \mathcal{J}$ & $\mathcal{F} \& \mathcal{J}$ & \textcolor{blue}{$\Delta$} & $\mathcal{F} \& \mathcal{J}$ & \textcolor{blue}{$\Delta$} \\ \midrule
LMPM & 37.2 & 22.6 & \textcolor{blue}{14.6} & 36.4 &  \textcolor{blue}{0.8} \\ 
VideoGLAMM & 48.2 & 21.6 & \textcolor{blue}{26.6} & 47.9 & \textcolor{blue}{0.3} \\ 
RGA &56.1 & 28.2 &  \textcolor{blue}{27.9} & 54.7 & \textcolor{blue}{1.4}\\
Sa2VA & 58.9 &  28.5 & \textcolor{blue}{30.4} & 54.8 & \textcolor{blue}{4.1}\\ \bottomrule
%MLLM + SAM 2.0 & 52.0 & 26.5 & \textcolor{blue}{25.5} & 47.2 & \textcolor{blue}{4.8}  \\
%MLLM + SAM 2.0 $\dagger$ & 57.4 & 28.1 & \textcolor{blue}{29.3} & 51.8 & \textcolor{blue}{5.6} \\ 
%Sa2VA$\star$ & \textbf{60.5} & \textbf{31.1} & \textcolor{blue}{29.4} & \textbf{57.5}  & \textcolor{blue}{3.0} \\  \bottomrule 
%Sa2VA$\star\star$ & \textbf{58.1} & \textbf{54.8} & \textcolor{blue}{3.3} & \textbf{55.6}  & \textcolor{blue}{2.5} \\
% Take time to evaluate this further and add the remaining results on training data ablation that clearly showed Sa2VA\star performance without any degradation almost on MoCentric-Bench beyond spatial biases confirmed
\end{tabular}}
\label{table:black_flag}
\end{table}

\begin{table}[t]
\centering
\caption{\textbf{Ablation of Motion Centric adaptation when averaging the predictions} of the two layouts $l_2$ and $l_3$ on our MeVIS \textit{val$^u$} variants.}
\resizebox{0.75\textwidth}{!}{
\begin{tabular}{l|c|ccc|c|ccc}
\toprule
Method & \textit{val$^u$} &  \multicolumn{3}{c|}{ \textit{val$^u$ \& Single frame}} &\textit{Reverse} & \multicolumn{3}{c}{\textit{val$^u$ \& Reverse} }  \\
 & $\mathcal{J\&F}$ &  $\mathcal{J}$  & $\mathcal{F}$ & $\mathcal{J\&F}$ & $\mathcal{J\&F}$ & $\mathcal{J}$ & $\mathcal{F}$ & $\mathcal{J\&F}$\\ \midrule
Sa2VA &58.9 & 25.5 & 38.2  & 31.8  & 61.1 &  26.0 & 38.7 & 32.4  \\ % Flip Results
%Sa2VA$\star$ - Run 1 &  26.8 & 39.2  & 33.0  &  28.5 & 42.4 & 35.4  \\ % Flip Results Run1
%Sa2VA$\star$  - Run 2 &  26.4 & 38.9  & 32.7  &  28.2 & 42.2 & 35.2  \\ % Flip Results Run2
Sa2VA$\star$ & \textbf{60.5} & \textbf{26.6} & \textbf{39.1}  &  \textbf{32.9} & \textbf{65.1} & \textbf{28.4} & \textbf{42.3} & \textbf{35.3}  \\ \bottomrule
%Sa2VA My Base - F & - &- & - & 26.3 & 39.0 & 32.6 &  - & - & - & 25.1 & 37.5 & 31.3 \\ % Flip Result for my Baseline
%Sa2VA$\dagger$ - F & - &- & - & 26.6 & 39.2  & 32.9 &  - & - & - & 26.4 & 40.0 & 33.2 \\ % Flip results and training with left and right random
%Sa2VA$\star$ - F & - &- & - & 26.6 & 39.1  & 32.9 &  - & - & - & 28.4 & 42.3 & 35.3 \\ \bottomrule % Flip Results avg
%Sa2VA$\star$ - Keyframe - F & ours (2025) & &  &  & 26.3 & 39.3 & 32.8 &  &  &   & 27.3  & 41.3 & 34.3 \\ \midrule
\end{tabular}}
\label{table:flip}
\end{table}

\begin{figure*}[t]
\centering
\begin{subfigure}{0.3\textwidth}
\includegraphics[width=\textwidth]{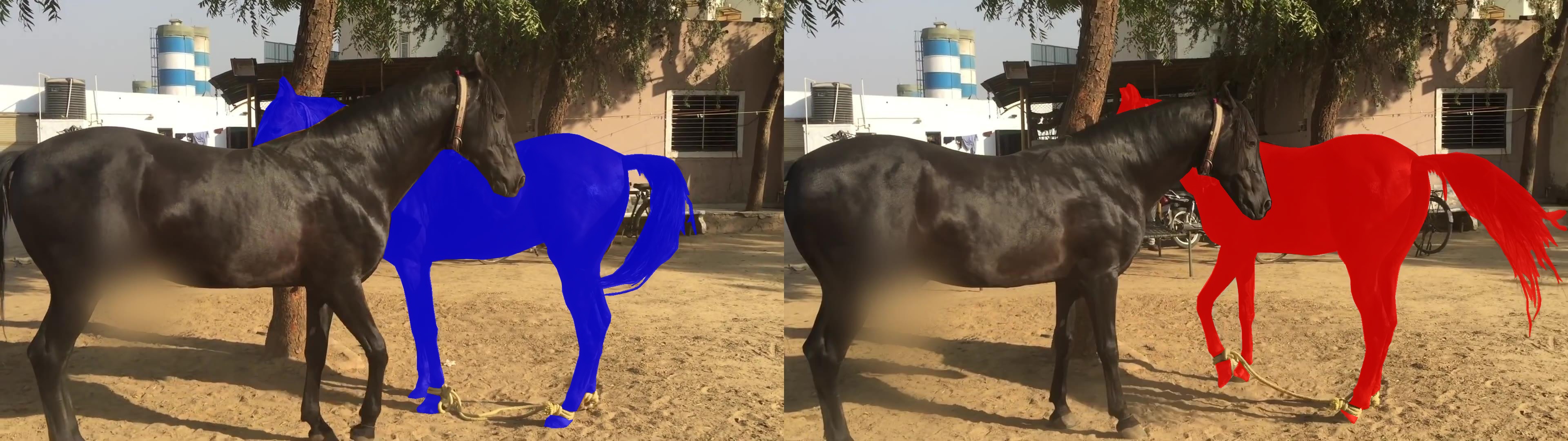}
\end{subfigure}%
\hspace{0.2em}%
\begin{subfigure}{0.3\textwidth}
\includegraphics[width=\textwidth]{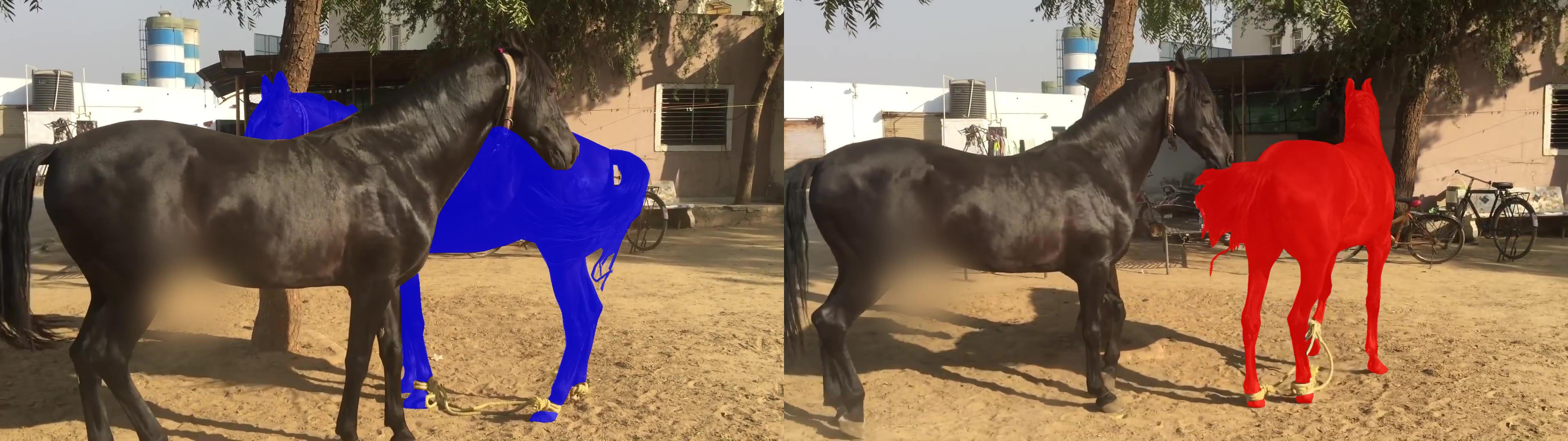}
\end{subfigure}%
\hspace{0.2em}%
\begin{subfigure}{0.3\textwidth}
\includegraphics[width=\textwidth]{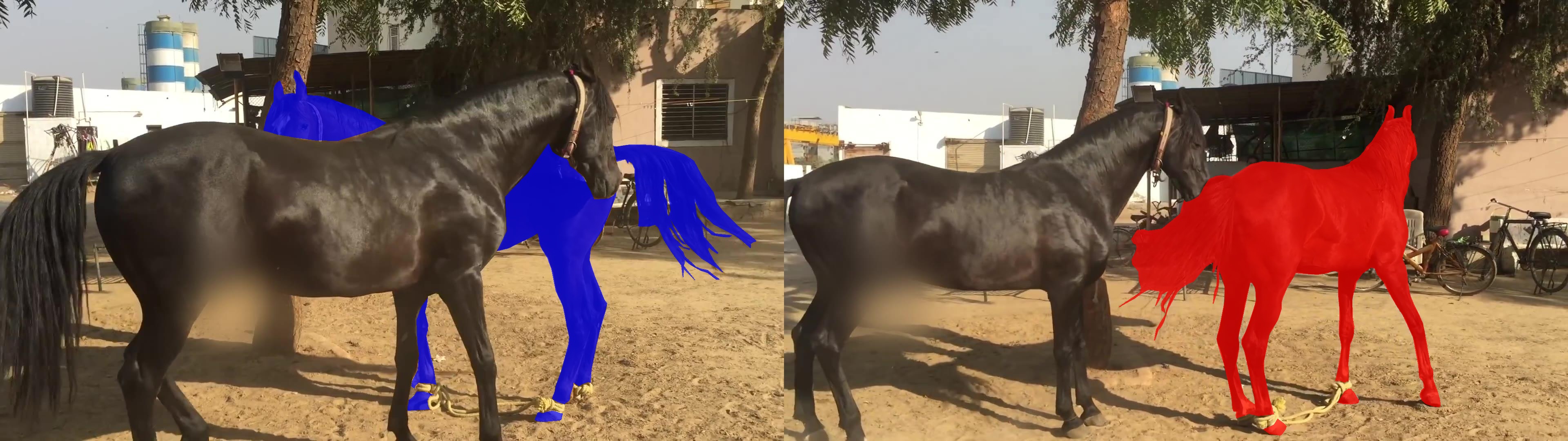}
\end{subfigure}

\begin{subfigure}{0.3\textwidth}
\includegraphics[width=\textwidth]{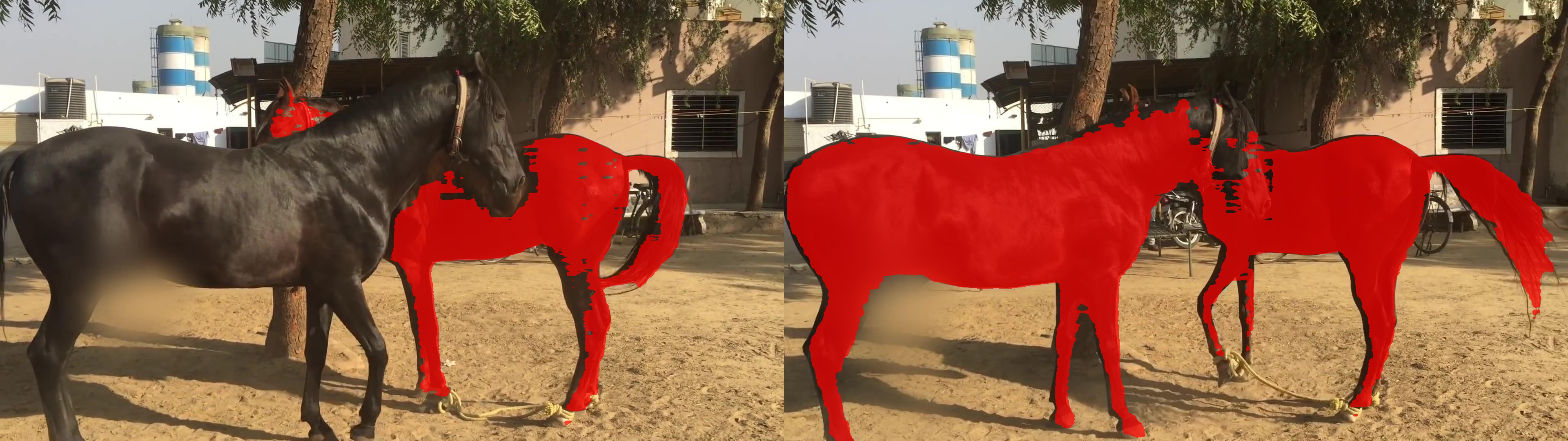}
\caption*{$I_1$}
\end{subfigure}%
\hspace{0.2em}%
\begin{subfigure}{0.3\textwidth}
\includegraphics[width=\textwidth]{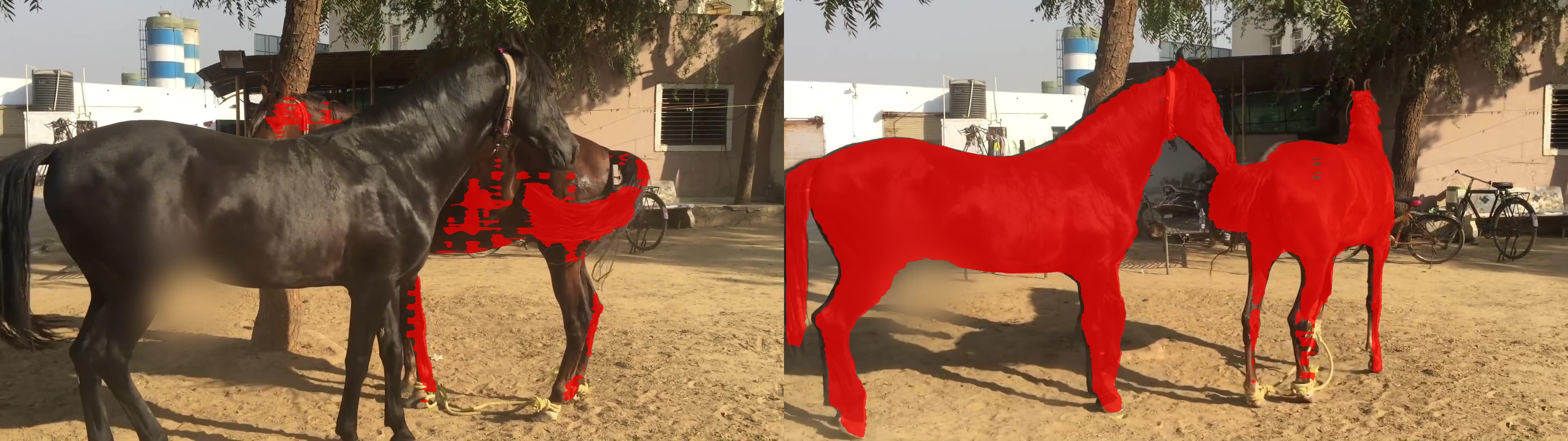}
\caption*{$I_2$}
\end{subfigure}%
\hspace{0.2em}%
\begin{subfigure}{0.3\textwidth}
\includegraphics[width=\textwidth]{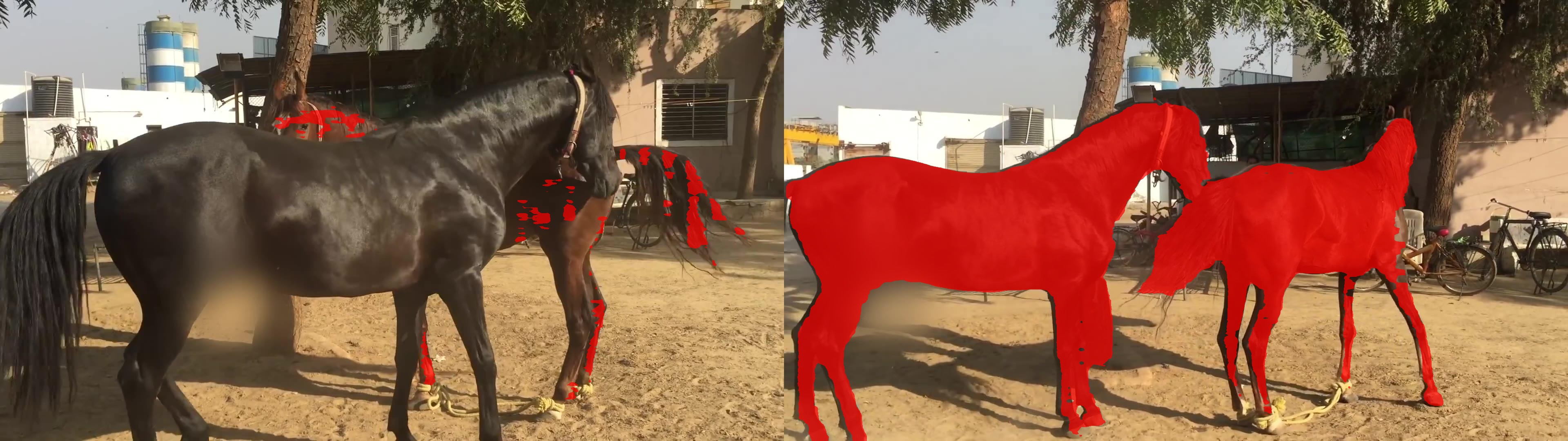}
\caption*{$I_3$}
\end{subfigure}
\caption{\textbf{Failure cases} of our motion-centric adaptation on the MeVIS variant \textit{val$^u$ \& Reverse}, the corresponding referred expression is ``the horse that starts with its head facing left and then turns to the right''. The frames shown correspond to frames 45, 55 and 60 near the end of the original video. \textbf{Top:} Groundtruth for the correct region highlighted in red and the deceiving region highlighted in blue. \textbf{Bottom:} The predictions of our motion centric adaptation of Sa2VA are highlighted in red.}
\label{fig:failures}
\end{figure*}

\textbf{Failure cases for our motion-centric adaptation.} Finally, we show failure cases of our motion-centric adaptation in Figure~\ref{fig:failures}, which mostly stem from underlying limitations in our base model, Sa2VA. Since it is limited to only the first five frames when identifying the motion expression. While our adaptation exhibits failures, we believe our benchmark and probing can push video MLLMs for visual grounding to have a proper understanding of local motion patterns beyond shortcuts that such models tend to perform when learning the interaction of motion and language. Examples of such shortcuts include the use of the heading or the object pose and the location, which we have observed to be tied with failures in the static keyframe in our referring expressions analysis.

\begin{figure*}[t]
\begin{minipage}{0.48\textwidth}
\captionof{table}{\textbf{Ablation study on the frame selection} mechanism in our single-image baselines on MeVIS \textit{val$^u$}.}
\centering
\resizebox{0.8\textwidth}{!}{
\begin{tabular}{l|ccc}
\toprule
 Method & \multicolumn{3}{c}{\textit{val$^u$}} \\
 &  $\mathcal{J}$ & $\mathcal{F}$ & $\mathcal{J\&F}$ \\ \midrule
First Frame & 47.9 & 56.2 & 52.0  \\ 
Last Frame & 49.2 & 57.5 & 53.3 \\ 
 KeyFrame & \textbf{54.0} & \textbf{60.9} & \textbf{57.4} \\  \toprule
\end{tabular}}
%\label{table:selection}
%\caption{Ablation study on the frame selection.}
\label{table:app_ablations}
\end{minipage}%
\hfill%
\begin{minipage}{0.48\textwidth}
\centering
\captionof{table}{\textbf{Ablation removing static expressions.} Sa2VA results confirm severe degradation in val$^u$ to Single frame \& val$^u$, even after excluding static expressions in the first probe, $\mathcal{P}_1$.}
\begin{tabular}{l|ccc}
\toprule
 Dataset & $\mathcal{J}$ & $\mathcal{F}$ & $\mathcal{J\&F}$\\ \midrule
\textit{val$^u$} & 56.4  & 64.8 & 60.6 \\ 
\textit{val$^u$ \& Single frame} & 25.1 & 35.9 & 30.5 \\ \bottomrule
%Single frame \& val$^u$ & 26.1 & 36.9 & 31.5 \\ \bottomrule %Sa2VA star
\end{tabular}
\label{table:nostatic}
\end{minipage}
\end{figure*}

\section{Limitations and Future Work}
\label{sec:limitations}
Aside from the failure cases that we observed in our motion-centric adaptation, we are currently confined to evaluating visual grounding on the pixel-level, which either discards models that are capable of bounding box/point level grounding only, or evaluates them with an additional component such as SAM 2.0. For our future work, we plan to expand our benchmark to include evaluation on the bounding box/point level to avoid this limitation. Furthermore, our benchmark relies on probing with multi-video layouts, which can affect models that capture motion by estimating the camera pose and geometry. Nonetheless, multiple videos on the internet, e.g., YouTube, exhibit such split-screen or multi-video layouts, which are part of standard video MLLMs benchmarks. Hence, this should be handled by such models that rely on geometric information. As part of our future work, we plan to provide a region-level captioning benchmark on our MoCentric-Bench. Due to the flexibility and extensibility of our probing, it is directly applicable in the aforementioned setting, which emphasizes the strength of our contribution.

\section{Impact Statement}
\label{sec:impact}
Video multi-modal large language models are widely used in various applications, such as robotics, human-computer interaction and medical image processing. The pixel-level understanding within such MLLMs is necessary for these applications that require the localization and, in certain scenarios, the delineation of the boundaries for the objects of interest. It is even more important to maintain a good spatiotemporal understanding to capture motion and dynamics in the input video. In our work, we have investigated the shortcomings of video MLLMs in the video referring segmentation task, while providing a more challenging motion-centric benchmark to push these models into a better understanding of motion. This is a significant development for AI safety to avoid the pitfalls and limitations of video MLLMs in capturing motion.

However, as with many other AI advancements, some risks could be entailed from the deployment of such models. There could be inherent biases emerging in such video  MLLMs, impacting various underrepresented groups. Nonetheless, studying these biases is out of the current scope of our work, yet we believe that our benchmarking efforts, probing and providing a tool to understand the pitfalls in the understanding and reasoning of these models are beneficial. We also believe, to the best of our knowledge, our introduced probing and evaluation does not introduce biases of its own towards such underrepresented groups. We leave it for future work to explore this further.

%%%%%%%%%%%%%%%%%%%%%%%%%%%%%%%%%%%%%%%%%%%%%%%%%%%%%%%%%%%%

%\newpage
%\input{checklist.tex}

\end{document}